\documentclass[10pt,twocolumn,letterpaper]{article}

\usepackage{iccv}
\usepackage{times}
\usepackage{epsfig}
\usepackage{graphicx}
\usepackage{amsmath}
\usepackage{amssymb}

\usepackage{booktabs}       
\usepackage{amsfonts}       
\usepackage{microtype}      
\usepackage{enumitem}
\usepackage{bm}
\usepackage{chngpage}
\usepackage{algorithm}
\usepackage{algorithmic}
\usepackage{subcaption}
\usepackage{wrapfig}
\usepackage{url}            

\usepackage{enumitem}
\setlist[enumerate]{itemsep=0mm}

\usepackage{nicefrac}       

\usepackage{color, colortbl}

\usepackage{adjustbox}
\usepackage{multirow}
\usepackage[english]{babel}

\newcommand{\bA}{\mathbf{A}}
\newcommand{\bJ}{\mathbf{J}}

\newcommand{\bx}{\mathbf{x}}
\newcommand{\by}{\mathbf{y}}
\newcommand{\bz}{\mathbf{z}}
\newcommand{\bbm}{\mathbf{m}}

\newcommand{\bbf}{\mathbf{f}}

\newcommand{\bI}{\mathbf{I}}

\newcommand{\bX}{\mathbf{X}}

\newcommand{\bxi}{\bm{\xi}}

\newcommand{\btheta}{{\bm{\theta}}}

\DeclareMathOperator{\Tr}{Tr}
\DeclareMathOperator{\Var}{Var}

\newtheorem{definition}{Definition}

\usepackage[pagebackref=true,breaklinks=true,letterpaper=true,colorlinks,bookmarks=false]{hyperref}

\iccvfinalcopy 

\def\iccvPaperID{3113} 

\ificcvfinal\fi
\begin{document}
	
	\title{A Simple yet Effective Baseline for Robust Deep Learning with Noisy Labels
	}
	
	\author{Yucen Luo	\thanks{Work done during an internship in Google Cloud AI.}
\\
	Tsinghua University\\
		{\tt\small luoyc15@mails.tsinghua.edu.cn}
		\and
		Jun Zhu\\
	    Tsinghua University\\
		{\tt\small dcszj@mail.tsinghua.edu.cn}
				\and
		Tomas Pfister\\
	    Google Cloud AI\\
		{\tt\small tpfister@google.com}
	}
	
	\maketitle

	\begin{abstract}
		Recently deep neural networks have shown their capacity to memorize training data, even with noisy labels, which hurts generalization performance. To mitigate this issue, we provide a simple but effective baseline method that is robust to noisy labels, even with severe noise. Our objective involves a variance regularization term that implicitly penalizes the Jacobian norm of the neural network on the whole training set (including the noisy-labeled data), which encourages generalization and prevents overfitting to the corrupted labels. Experiments on both synthetically generated incorrect labels and realistic large-scale noisy datasets demonstrate that our approach achieves state-of-the-art performance with a high tolerance to severe noise.
	\end{abstract}
	
	\section{Introduction}
	
	Recently deep neural networks (DNNs) have achieved remarkable performance on many tasks, such as speech recognition~\cite{amodei2016deep}, image classification~\cite{he2016deep}, object detection~\cite{ren2015faster}. 
	However, DNNs usually need a large-scale training dataset to generalize well. 
	Such large-scale datasets can be collected by crowd-sourcing, web crawling and machine generation with a relative low price, but the labeling may contain errors. 
	Recent studies~\cite{zhang2016understanding, arpit2017closer} reveal that mislabeled examples hurt generalization. 
	Even worse, DNNs can memorize the training data with completely randomly-flipped labels, which indicates that DNNs are prone to overfit noisy training data. 
	Therefore, it is crucial to develop algorithms robust to various amounts of label noise that still obtain good generalization.

	To address the degraded generalization of training with noisy labels, one direct approach is to reweigh training examples~\cite{ren2018learning,jiang2017mentornet,han2018co,ma2018dimensionality}, which is related to curriculum learning. 
	The general idea is to assign important weights to examples with a high chance of being correct. 
	However, there are two major limitations of existing methods. 
	First, imagine an ideal weighting mechanism. 
	It will only focus on the selected clean examples. 
	For those incorrectly labeled data samples, the weights should be near zero. 
	If a dataset is under 80\% noise corruption, an ideal weighting mechanism assigns nonzero weights to only 20\% examples and abandons the information in a large amount of 80\% examples. 
	This leads to an insufficient usage of training data.  
	Second, previous methods usually need some prior knowledge on the noise ratio or the availability of an additional clean unbiased validation dataset. 
	But it is usually impractical to get this extra information in real applications. 
	Another approach is correction-based, \ie estimating the noisy corruption matrix and correcting the labels~\cite{patrini2017making,reed2014training, goldberger2016training}. 
	But it is often difficult to estimate the underlying noise corruption matrix when the number of classes is large. 
	Further, there may not be an underlying ground truth corruption process but an open set of noisy labels in the real world. 
	Although many complex approaches~\cite{jiang2017mentornet, ren2018learning,han2018co} have been proposed to deal with label noise, we find that a simple yet effective baseline can achieve surprisingly good performance compared to the strong competing methods.
	
	In this paper, 
	we propose to minimize the predictive variance, which is an unbiased estimator of Jacobian norm.
	A model with simpler hypothesis and smoother decision boundaries is assumed to generalize better.
	Our method is simple yet effective which can take advantage of the whole dataset including the noisy examples to improve the generalization. 
	
	Our main contributions are:
	\begin{itemize}
		\item We propose a new strong baseline method for robustness to noisy labels, which greatly mitigates over-fitting and should not be omitted in the label noise community. A thorough empirical evaluation on various datasets (\eg, CIFAR-10, CIFAR-100, ImageNet) is conducted and demonstrates significant improvements over previous competing methods. 
		We also apply our method to a large-scale real-world  noisy dataset, Webvision~\cite{li2017webvision}, and establish the new state-of-the-art results.
		
		\item We show that the variance-based regularizer is an unbiased estimator of Jacobian norm and analyze the reliability of this estimator. Its good performance is due to that Jacobian norm correlates with generalization.
		The method can be applied to any neural network architecture. Additional knowledge on the clean validation dataset is not required.
		
		\item Empirically we find that our method learns a model with lower subspace dimensionality and lower complexity, which are the indicators of better generalization.
	\end{itemize}

	\section{Related work}
	\label{related}
	Learning with noisy labels has been broadly studied in previous work, both theoretically~\cite{natarajan2013learning} and empirically~\cite{reed2014training,han2018co,jiang2017mentornet}. Here we focus on the recent progress on deep learning with noisy labels. Since DNNs have high capacity to fit the (noisy) data, it brings new challenges different from that in the traditional noisy label settings.
	
	\textbf{Generalization of DNNs.}
	Previous works~\cite{zhang2016understanding,arpit2017closer,shwartz2017opening} find that DNNs have different learning patterns for clean or noisy labels. \cite{zhang2016understanding} shows that DNNs can easily memorize the training dataset even when the labels are random noise. 
	An early stage of pattern learning and later memorization of noisy labels are observed in~\cite{arpit2017closer}.
	Some previous work uses this kind of property to propose measures to modify the training process, such as the learned subspace dimensionality and the distribution of the loss values~\cite{ma2018dimensionality,han2018co}. 
	Regularization techniques including dropout and early stopping have been shown to be effective to prevent over-fitting to noisy labels~\cite{arpit2017closer}.
	
	\textbf{Estimating noise distribution.} 
	Many noisy estimation models have been proposed~\cite{natarajan2013learning,xiao2015learning,patrini2017making,vahdat2017toward}. 
	Some works assume the true label is modeled by a latent variable while the noisy label is observed. 
	EM-like methods have been proposed to alternate between the learning of noisy corruption process and the modeling. 
	Backward and forward corrections~\cite{patrini2017making} use an estimated noise transition matrix to modify the loss function.
	In general, the noise is assumed to be input-independent but class-dependent. Input-dependent noise has been explored in~\cite{menon2016learning, xiao2015learning}.
	
	\textbf{Noise-robust loss functions.} 
	The mean absolute error (MAE) was proposed as a noise-robust alternative to the cross-entropy loss~\cite{ghosh2017robust} but was known to be hard to converge. 
	An extension and generalization of MAE, generalized cross entropy, was recently developed~\cite{zhang2018generalized}.
	
	\textbf{Identifying clean examples.} 
	Co-teaching~\cite{han2018co} proposes to identify the examples with small loss as clean examples. Learning to reweight~\cite{ren2018learning} equals to shifting the training distribution $p(x,y)$ to match the clean validation distribution $q(x,y)$, that is to minimize $D_{f_\theta}(w(x,y)p(x,y), q(x,y))$ where $D$ is some distance measure implicitly learned by $f_\theta$ and $w(x,y)$ is the density ratio, \ie the learned weights for each example $(x,y)$.
	
	\textbf{Using additional clean validation dataset.}
	\cite{azadi2015auxiliary} proposed a regularization term to encourage the model to select reliable examples.
	\cite{hendrycks2018using} proposed Golden Loss Correction to use a set of trusted clean data to mitigate the effects of label noise. They estimate the corruption matrix using the trained network with noisy labels and then re-train the network corrected by the corruption matrix. 
	\cite{ren2018learning} also used a small clean validation dataset to determine the weights of training examples.
	The success of these methods is based on the assumption that clean data is from the same distribution as the corrupted data as well as the test data. 
	However, more realistic scenario are ones where (1) $p(x)$ varies between the clean data and the noisy data, e.g., imbalanced datasets. 2) There is class mismatch: $p(y|x)$ differs. 
	Similar problems exist in semi-supervised learning.
	All these methods require a clean validation dataset to work well while the proposed method does not require it.

	\section{Preliminaries}
	In this section, we first briefly introduce some notations and settings and then provide a measure to assess the learning behaviour of DNNs during the training.
	
	The target is to learn a robust $K$-class classifier $f$ from a training dataset of images with noisy supervision. 
	Let $\mathcal{D} = \{(x_1, \tilde y_1), ..., (x_N, \tilde y_N)\}$ denote a training dataset, where $x_n\in X$ is the $n$-th image in sample space $X$ (e.g., $\mathbb{R}^d$) with its corresponding noisy label $\tilde y_n\in\{1, 2,..., K\}$.

	\subsection{Label noise}
	
	The label noise is often assumed to be class-conditional noise in previous work~\cite{natarajan2013learning,patrini2017making}, where the true label $y$ is flipped to $\tilde y\in \mathcal{Y}$ with some probability $p(\tilde y | y)$. It means that we assume $p(\tilde y | x, y) = p(\tilde y| y)$, \ie, the corruption of labels is independent of the input $x$. 
	This kind of assumption is an abstract approximation to the real-world corruption process.
	For example, non-expert labelers may fail to distinguish some specific species. 
	The probability $p(\tilde y | y)$ is represented by a noise transition matrix $T \in [0,1]^{K \times K}$, where $T_{ij}=p(\tilde y = j | y = i)$.
	The examples $(x_i, \tilde y_i)$ in $\mathcal{D}$ are sampled from $p(x,\tilde y) = \sum_y p(\tilde y | y) p(y|x) p(x)$, a distribution that marginalizes over the unknown true label.
	A few exceptions~\cite{xiao2015learning,menon2016learning} also consider the input-dependent noise model $p(\tilde y | x, y)$.
	
	\subsection{Local Intrinsic Dimensionality (LID)}
	\label{sec:lid}
	Local Intrinsic Dimensionality (LID) has been used to assess the learning behavior of DNNs~\cite{ma2018dimensionality}. 
	The subspace dimensionality is affected by the quality of labels. When the model is learned with clean labels, the LID score decreases as the training proceeds. But a different behavior is observed when the model is learned with noisy labels. A poorly-regularized deep network tends to increase the LID score after an initial decrease, which corresponds to the increased test accuracy followed by a decrease.
	Formally, LID is defined in~\cite{houle2017local,ma2018dimensionality}:
	
\begin{definition}[Local Intrinsic Dimensionality] \quad \\
	Given a data sample $x \in X$, let $r>0$ be a random variable denoting the distance from $x$ to other data samples. If the cumulative distribution function $F(r)$ is positive and continuously differentiable at distance $r>0$, the LID of $x$ at distance $r$ is given by:
	\begin{equation} \label{eq:LID_r}
	\begin{split}
	\textup{LID}_F(r) & \triangleq \lim_{\epsilon\to 0} \frac{\ln\big(F((1+\epsilon) r)\big/F(r)\big)}{\ln(1+\epsilon)} 
	= \frac{r F'(r)}{F(r)},
	\end{split}
	\end{equation}
	whenever the limit exists.
	\label{def:lid}
	The LID at $x$ is in turn defined as the limit of $r \to 0$: 
	\begin{equation} \label{eq:LID}
	\textup{LID}_F = \lim_{r \to 0}  \textup{LID}_F(r).
	\end{equation}
\end{definition}

LID gives an indication of the dimension of sub-manifold including $x$ locally. In practice, many estimators of LID have been proposed, among which Maximum Likelihood Estimator (MLE) can trade-off between the efficiency and complexity:
\begin{equation} \label{eq:estimator}
\widehat{\textup{LID}}(x) = - \Bigg( \frac{1}{k}\sum_{i=1}^{k}\log \frac{r_i(x)}{r_{\mathit{max}}(x)}\Bigg)^{-1},
\end{equation}
where $r_i(x)$ is the distance between $x$ and its $i$-th nearest neighboring point, and $r_{\mathit{max}}(x)$ denotes the maximum of distances to all the neighbors.

Further, as it is computationally expensive to get the LID estimator by calculating the distance within the whole dataset, a stochastic mini-batch version is proposed in~\cite{ma2018dimensionality}. Given a batch of training data $X_B$, the LID score at $x$ can be estimated as:
\begin{equation} \label{eq:lid_training}
\widehat{\textup{LID}}(x, X_B)\! =\! - \Bigg( \frac{1}{k}\sum_{i=1}^{k}\log \frac{r_i(g(x), g(X_{B}))}{r_{\mathit{max}}(g(x), g(X_{B}))}\Bigg)^{-1},
\end{equation}
where $g(x)$ is the feature learned by DNNs.
$\widehat{\text{LID}}(x, X_B)$ indicates the dimensional complexity in the vicinity of $x$. In practice, the approximation is accurate enough if we take a large batch size. We will use this measure later to assess the degree of over-fitting in the training process. In general, a lower LID score indicates a better generalization~\cite{ma2018dimensionality}.

\section{Our approach}
	

	In this section, we present a robust training algorithm to deal with noisy labels. Since DNNs tend to increase its complexity and LID to accommodate the noisy labels but have poor generalization, we argue that a model with lower complexity would be more robust to label noise and generalizes better. 
	The dimensionality of the learned subspace and the smoothness of decision boundaries can both indicate how complex the model is. Therefore, we propose a method to regularize the predictive variance to achieve low subspace dimensionality and smoothness.

	\subsection{Variance-based regularization}
	In order to alleviate over-fitting to the label noise, 
	we propose a regularizer that is not dependent on the labels. 
	We induce the smoothness of decision boundaries along the data manifold, which is shown to improve the generalization and robustness.
	If an example $x$ is incorrectly labeled with $\tilde y$, it has a high probability to lie near the decision boundary or in the wrong cluster not belonging to $y$. 
	Therefore, the prediction variance can be high on the noisy examples. 
	We propose to regularize the variance term. 
	The mapping function is smoothed and thus also the decision boundaries.
	
	Concretely, the variance is estimated by the difference of predictions under perturbations $\bxi$ and $\bxi'$:
	\begin{eqnarray}
	\label{equ:3}
R_V(\bX, \btheta) = \frac{1}{N}\sum_{i=1}^{N}\mathbb{E}_{\bxi',\bxi} \;\| \bbf(\bx_i;\btheta, \bxi') - \bbf(\bx_i;\btheta, \bxi)\|^2,
	\end{eqnarray}
	where $\bX$ is the training set and $\theta$ denotes the network parameter.
	We show that $R_V(\bX, \btheta)$ is an unbiased estimation of the predictive variance if the input perturbations are also treated as a part of the model uncertainty.
	
\begin{align}
	R_V(\bX, \btheta) &= \frac{1}{N}\sum_{i=1}^{N}\mathbb{E}_{\bxi',\bxi}\| \bbf(\bx_i;\btheta, \bxi') - \bbf(\bx_i;\btheta, \bxi)\|^2\notag\\
	& = \frac{1}{N}\sum_{i=1}^{N} \Big[\mathbb{E}_{\bxi'}\|\bbf(\bx_i; \btheta, \bxi')\|^2  +  \mathbb{E}_{\bxi}\|\bbf(\bx_i; \btheta, \bxi)\|^2\notag\\
	&\quad\; - 2\mathbb{E}_{\bxi'}\bbf(\bx_i; \btheta, \bxi')^\top \mathbb{E}_{\bxi}\bbf(\bx_i; \btheta, \bxi)\Big]\label{eq1}\\
	&= \frac{1}{N}\sum_{i=1}^{N}\Big[2\mathbb{E}_{\bxi}\|\bbf(\bx_i; \btheta, \bxi)\|^2 -2\|\mathbb{E}_{\bxi}\bbf(\bx_i;\btheta, \bxi)\|^2\Big]\label{eq2}\\
	&=\frac{2}{N}\sum_{i=1}^{N} \sum_{k=1}^{K}\mathrm{Var}_{\bxi}[\bbf(\bx_i;\btheta, \bxi)]_k\label{rv}
\end{align}
where the input $\bx_i\in \mathbb{R}^{D}$ and the prediction function is $\bbf:\mathbb{R}^{D} \to \mathbb{R}^{K}$. The perturbations $\bxi,\bxi'$ are assumed to be i.i.d. random variables including input Gaussian noise, random data augmentation and network noise like dropout. From Eq.~\ref{eq1} to Eq.~\ref{eq2}, we use the property of i.i.d. r.v. $\bxi,\bxi'$, \ie $\mathbb{E}_{\bxi'}\bbf(\bx_i; \btheta, \bxi') = \mathbb{E}_{\bxi}\bbf(\bx_i; \btheta, \bxi)$ and $\mathbb{E}_{\bxi'}\|\bbf(\bx_i; \btheta, \bxi')\|^2 = \mathbb{E}_{\bxi}\|\bbf(\bx_i; \btheta, \bxi)\|^2$.
In practice, the expectation is implemented by stochastic samples of the perturbations $\bxi_i,\bxi'_i$, \ie, we use 
\begin{align}
\label{rvhat}
    \hat R_V(\bX, \btheta) = \frac{1}{N}\sum_{i=1}^{N}\| \bbf(\bx_i;\btheta, \bxi_i') - \bbf(\bx_i;\btheta, \bxi_i)\|^2,
\end{align}
and easily $R_V(\bX, \btheta)$ is the expectation of $\hat R_V(\bX, \btheta)$ over perturbations, \ie, $\mathbb{E}_{\bxi, \bxi'}\hat R_V(\bX, \btheta) = R_V(\bX, \btheta)$.
From Eq.~\ref{rv} we can see that $R_V$ is equivalent to the sum of variance of the prediction each dimension under some perturbations.
	
	\textbf{Relation to the generalization of DNNs.}
	We show that this regularization helps to learn a low-dimensional feature space that captures the underlying data distribution. 
	The variance term implicitly estimates the Jacobian norm of the neural network: 
	\begin{align}
	    	\|\bJ(\bx)\|_F .
	\end{align}
	
	A simplified version to analyze is to assume $\bxi, \bxi'$ are i.i.d. sampled from a Gaussian distribution, i.e., $\bxi, \bxi' \stackrel{\text{i.i.d.}}{\sim} \mathcal{N}(\bm0,\sigma^2 \bI_D)$ and the perturbations are infinitesimal and additive, \ie, $\tilde \bx = \bx + \bxi$ where $\sigma$ is near zero.
	\begin{align}
	&\frac{1}{\sigma^2}R_V(\bX, \btheta)\notag\\
	&= \frac{1}{\sigma^2} \frac{1}{N}\sum_{i=1}^{N}\mathbb{E}_{\bxi',\bxi}\|\bbf(\bx_i + \bxi;\btheta) - \bbf(\bx_i + \bxi';\btheta)\|^2
	\end{align}
	 By first-order Taylor expansion, and let $\bJ(\bx) = \frac{\partial \bbf}{\partial \bx} \in \mathbb{R}^{K\times D}$
\begin{align}
    \bbf(\bx+\bxi) = \bbf(\bx) + \bJ(\bx) \bxi + o(\bxi),
    \end{align}
    and omitting the high-order terms, we have
    \begin{align}
    &\frac{1}{\sigma^2}R_V(\bX, \btheta)\notag\\
    =&\frac{1}{\sigma^2}\frac{1}{N}\sum_{i=1}^{N} \mathbb{E}_{\bxi}\|\bJ(\bx_i)\bxi\|^2 +\mathbb{E}_{\bxi'}\|\bJ(\bx_i)\bxi'\|^2 \notag\\
    &\quad- 2 \mathbb{E}_{\bxi}\bxi^\top \bJ(\bx_i)^\top \bJ(\bx_i) \mathbb{E}_{\bxi}\bxi'\notag\\
    =&2 \frac{1}{\sigma^2}\frac{1}{N}\sum_{i=1}^{N} \mathbb{E}_{\bxi}\|\bJ(\bx_i)\bxi\|^2 - [\|\mathbb{E}_{\bxi} \bJ(\bx_i)\bxi\|]^2 \notag\\
    =& 2 \frac{1}{\sigma^2}\frac{1}{N}\sum_{i=1}^{N} \mathbb{E}_{\bxi}[\bxi^\top \bJ(\bx_i)^\top \bJ(\bx_i)\bxi] - 0\label{eq6}\\
    =&2\frac{1}{N}\sum_{i=1}^{N} \Tr\left[\bJ(\bx_i)^\top \bJ(\bx_i)
    \frac{1}{\sigma^2} \mathbb{E}_{\bxi} [\bxi\bxi^\top ]\right]\\
    =&2\frac{1}{N}\sum_{i=1}^{N} \Tr\left[\bJ(\bx_i)^\top \bJ(\bx_i)\frac{1}{\sigma^2} \sigma^2 \bI_D\right] \label{eq8}\\
    =&2\frac{1}{N}\sum_{i=1}^{N} \|\bJ(\bx_i)\|_F^2. 
    \end{align}
    If we further take expectation over $N$ samples of $\bx_i$, we get
    \begin{align}
      \mathbb{E}_{\bX}\frac{1}{\sigma^2}R_V(\bX, \btheta) =  2 \frac{1}{N}N\mathbb{E}_\bx  \|\bJ(\bx)\|_F^2=  2 \mathbb{E}_\bx  \|\bJ(\bx)\|_F^2.
    \end{align}
The we can prove that $\frac{1}{2\sigma^2}\hat R_V(\bX, \btheta)$
is an unbiased estimator of $\mathbb{E}_{\bx}  \|\bJ(\bx)\|_F^2$ using the fact that $\mathbb{E}_{\bxi, \bxi'}\hat R_V(\bX, \btheta) = R_V(\bX, \btheta)$, \ie, 
\begin{align}
\mathbb{E}_{\bX,\bxi}\frac{1}{2\sigma^2}\hat R_V(\bX, \btheta)= \mathbb{E}_\bx  \|\bJ(\bx)\|_F^2.
\end{align}
The expectation of the estimator is the Jacobian norm.
Theoretically it was pointed out in~\cite{sokolic2017robust} that a bounded spectral norm of the network's Jacobian matrix in the neighbourhood of the training examples is crucial for DNNs to generalize well. Empirically, \cite{novak2018sensitivity} observed that the change in generalization is coupled with the respective change in sensitivity as measured by the Jacobian norm, \ie, lower sensitivity corresponds to smaller generalization gap.

And the next question is how reliable the estimator $\frac{1}{2\sigma^2}\hat R_V(\bX, \btheta)$ is. What can we say about the variance of this estimator? We know that an estimator with low variance will give more confidence in its usage.

Let $\bA = \bJ(\bx)^\top \bJ(\bx)$ and $\bz$ be the normalized $\bxi$, \ie, $\bz = \frac{1}{\sigma} \bxi$ is a zero-mean, unit-variance random variable. From Eqs.~\ref{eq6}-\ref{eq8}, we can get
\begin{align}
  \Tr(\bA) = \mathbb{E}_\bz[\bz^\top \bA \bz]  
\end{align}
For simplicity, we first derive the variance of $\bz^\top \bA \bz$ and then the variance of $\frac{1}{2\sigma^2}\hat R_V(\bX, \btheta) = \mathbb{E}_\bx \mathbb{E}_\bz[\bz^\top \bA \bz]$ can be obtained by law of total variance.

Let $\mathbf{a} = diag(\mathbf{A})$, $\bbm = \mathbb{E}\bz = \bm0$,
\begin{align}
    \Var[\mathbf{z}^\top\mathbf{A}\mathbf{z}] = &2\mu_2^2 \Tr(\mathbf{A}\mathbf{A}^\top) + 4 \mu_2\mathbf{m}^\top\mathbf{A} \mathbf{m} + 4 \mu_3\mathbf{m}^\top\mathbf{A}\mathbf{a} \notag\\
    &+ (\mu_4 - 3 \mu_2^2)\mathbf{a}^\top\mathbf{a}
\end{align}
For normal Gaussian random variable $\bz$, the second central moment (variance) $\mu_2 = 1$, and $\mu_4 = 3$.
Then we have 
$$ \Var[\mathbf{z}^\top\mathbf{A}\mathbf{z}] = 2\|\mathbf{A}\|_F^2 = 2\|\bJ(\bx)^\top \bJ(\bx)\|_F^2.$$
For the estimator $\bz^\top \bA \bz$, a bound on the number of samples needed by the Monte Carlo estimator to obtain an error of at most $\epsilon$ with probability $\delta$ is $O(20 \epsilon^{-2} \ln \left(\frac{2}{\delta}\right))$.

By the law of total variance $\Var(\by) = \mathbb{E}[\Var(\by|\bx)] + \Var[\mathbb{E}(\by|\bx)]$,
here $\frac{1}{2\sigma^2}\hat R_V(\bX, \btheta)$ is treated as $\by$, we have
\begin{align}
    &\Var_{\bx, \bxi}[\frac{1}{2\sigma^2}\hat R_V(\bX, \btheta)] =\notag\\
    &2\mathbb{E}_\bx\|\bJ(\bx)^\top \bJ(\bx)\|_F^2 + \Var_\bx[\|\bJ(\bx)\|_F^2]].
\end{align}
Therefore, the variance of the our unbiased estimator to the Jacobian norm is bounded, which means it makes sense to use this estimator for the Jacobian norm.

Above we derive the case of random normal perturbations,
perturbations on the data manifold $\mathcal{M}$ can be approximated by stochastic data augmentation in reality. It is assumed that a natural image with data augmentation still lies on the manifold of images. It can be proved that the variance-based regularizer under stochastic data augmentation is equivalent to the Jacobian norm along the manifold $\bJ_\mathcal{M}$. The proof is similar as the above except that the random noise $\bxi$ is projected to the manifold, turning into $\bxi_\mathcal{M}$.

\textbf{Relation to semi-supervised learning.}
	Similar objectives called consistency loss have been successfully explored in semi-supervised learning~\cite{laine2016temporal,tarvainen2017mean} but with different motivations, which aim to leverage the unlabeled data while ours is from the perspective of resistance to label noise. Examples without labels in semi-supervised learning are not detrimental to the learning if i.i.d. assumption of data holds. However, in our setting, examples with wrong labels will hurt the learning process, especially the non-uniform noise. Our paper aims to empirically show that the simple objective provides a strong baseline for the robustness to label noise, which should not be omitted int the label noise community. We have also provided some insights on why it works well above, \ie, the regularizer is an unbiased estimator of the Jacobian norm with bounded variance. 
	
	\textbf{Relation to posterior regularization.}
    Minimizing the predictive variance has been applied to deep kernel learning on regression tasks~\cite{jean2018semi}. It was pointed out that variance minimization can be explained in the framework of posterior regularization. Optimizing the objective is equivalent to computing a regularized posterior by solving a regularized Bayesian inference (RegBayes) optimization problem~\cite{zhu2014bayesian,jean2018semi}. It restricts the solution to be of some specific form, which is equivalent to imposing some prior knowledge of the model structure.
    The regularizer serves as an inductive bias on the structure of the feature space. By reducing the variance of predictions, the neural network is encouraged to learn a low-dimensional feature space where the training examples are far from the decision boundaries and tend to cluster together (see Figs~\ref{lid}, \ref{csr}, \ref{visual} for empirical results). This alleviates the possibility of the model to increase its complexity to fit the noisy labels.
	
	Therefore, the learning objective is simply
	\begin{equation}
	\label{equ:2}
	\min_{\btheta}\sum\limits_{i=1}^N\ell(\bbf(\bx_i;\btheta), \tilde y_i) + \lambda \hat R_V(\bX, \btheta),
\end{equation}
where the first term is any supervised loss function including the cross-entropy loss or previously proposed noise-robust losses. In Section~\ref{exp}, we show empirically that the objective can learn a model with low subspace dimensionality and low hypothesis complexity. It is noted that the methods mentioned in Section~\ref{related} including filtering out noisy examples and robust supervised losses are orthogonal to our approach and the combinations can yield better performance. But to focus on this simple baseline, we only provide results without any other modifications.

		\begin{table*}[t]
		\caption{Averaged test error rates (\%) and the standard deviations over 3 runs on CIFAR-10 under different uniform noise fraction. Methods marked with $^\dagger$ are trained using additional clean validation images. Best results are highlighted in bold.}
		\label{cifar}
		\begin{adjustbox}{width=\textwidth}
	\small
			\begin{tabular}{lllllll}
				\toprule
				\multirow{2}{*}
				{Methods} &\multicolumn{5}{c}{Noise Ratio $\eta$}& \multirow{2}{*}{Network} \\ \cline{2-6} 
				
				&0    &0.2   &0.4   &0.6 &0.8 &  \\
				\midrule
				Bootstrap-hard~\cite{reed2014training} & 10.94 $\pm$ 0.9& 20.81 $\pm$ 0.4 & 23.33 $\pm$ 0.8& 29.43 $\pm$ 0.3& --&12-layer CNN\\
				\midrule
				Forward-correction~\cite{patrini2017making} &9.73 $\pm$ 0.0& 15.39 $\pm$ 0.3& 18.16 $\pm$ 0.1& 27.59 $\pm$ 0.7& --&12-layer CNN\\
				\midrule
				D2L~\cite{ma2018dimensionality}& 10.59  $\pm$  0.2& 14.87  $\pm$  0.6& 16.64 $\pm$  0.5& 27.16 $\pm$  0.6&-- &12-layer CNN\\
				\midrule
				Generalized Cross Entropy~\cite{zhang2018generalized} & 6.5 &10.13  $\pm$  0.2 &12.87  $\pm$  0.22& 17.46  $\pm$  0.23 &32.08  $\pm$  0.6& ResNet-34\\
				\midrule
				
				Co-teaching~\cite{han2018co}&6.05 &17.68&-- & --&-- & 13-layer CNN\\
				\midrule
				
				MentorNet~\cite{jiang2017mentornet}$^\dagger$	& 4 & 8  & 11 &-- & 51 & WRN-101-10\\
				\midrule
				Learning to reweight~\cite{ren2018learning}$^\dagger$& 3.87 	 		& --			& 13.08 $\pm$ 0.19&--& --&WRN-28-10\\
												\midrule

					Ours& {3.79  $\pm$  0.13} &\textbf{ 3.87  $\pm$  0.15}		&  \textbf{5.05 $\pm$  0.24} &\textbf{6.42$\pm$  0.28} & \textbf{13.31 $\pm$  0.45}&WRN-28-10\\
				\bottomrule
			\end{tabular}
		\end{adjustbox}
	\end{table*}
	\begin{table*}[t]
		\caption{Test error rates (\%) on CIFAR-100 under different uniform noise fraction. Methods marked with $^\dagger$ are trained using additional clean validation images. Best results are highlighted in bold.}
		\label{cifar100}
		\centering
	\begin{adjustbox}{width=\textwidth}
	\small
			\begin{tabular}{lllllll}
				\toprule
				\multirow{2}{*}
				{Methods} &\multicolumn{5}{c}{Noise Ratio $\eta$}& \multirow{2}{*}{Network} \\ \cline{2-6} 
				
				&0    &0.2   &0.4   &0.6 &0.8 &  \\
				\midrule
				Bootstrap-hard~\cite{reed2014training} & 31.69 $\pm$ 0.2& 41.51 $\pm$ 0.4& 53.56 $\pm$ 0.7& 57.35 $\pm$ 0.9& -- & ResNet-44
				\\
				\midrule
				Forward-correction~\cite{patrini2017making} & 31.46 $\pm$ 0.1 & 39.75 $\pm$ 0.2
				 & 48.73 $\pm$ 0.3& 55.78 $\pm$  0.7& -- &ResNet-44
				\\\midrule
				D2L~\cite{ma2018dimensionality}& 31.40 $\pm$ 0.3& 37.80  $\pm$  0.5& 46.99 $\pm$ 0.7& 54.79  $\pm$  0.4 & -- & ResNet-44
				\\\midrule
				Generalized Cross Entropy~\cite{zhang2018generalized} & 28.6 &33.19 $\pm$ 0.42 & 38.23 $\pm$ 0.24 & 45.96 $\pm$  0.56 &52.34 $\pm$ 0.69& ResNet-34
				\\\midrule
				Co-teaching~\cite{han2018co}& 29.15 &45.77&--  & --& -- & 13-layer CNN
				\\\midrule

				MentorNet~\cite{jiang2017mentornet}$^\dagger$	& 21 &  27  & 32&-- & 65& WRN-101-10\\
				\midrule
				Learning to reweight~\cite{ren2018learning}$^\dagger$& 21.8	 		& --			& 38.66 $\pm$ 2.06 &-- & --& WRN-28-10\\
				\midrule
				Ours& {18.6$\pm$0.15} & \textbf{19.45  $\pm$0.22}			&  \textbf{25.73  $\pm$  0.47}  & \textbf{38.23  $\pm$0.52} & \textbf{44.68 $\pm$ 0.75}& WRN-28-10\\
				\bottomrule
			\end{tabular}
			\end{adjustbox}
	\end{table*}
	
		\begin{table*}[t]\vspace{-0.3in}
		\caption{Results on CIFAR-10 and CIFAR-100 with class-dependent asymmetric noise. Averaged accuracy and standard deviation over 3 runs are reported. The results of competing methods are taken from~\cite{zhang2018generalized}. CCE stands for commonly-used categorical cross-entropy loss function, MAE stands for mean absolute error. Forward $T^\dagger$~\cite{patrini2017making} uses the ground-truth noise transition matrix while Forward $\hat T$~\cite{patrini2017making} estimates $T$. Comparison to Forward $T^\dagger$ is not fair. Trunc $\mathcal{L}_q$ loss is a noise-robust loss function proposed in~\cite{zhang2018generalized}.}
				\label{classdep}
		\centering
		\begin{adjustbox}{width=0.8\textwidth}
			\small
			\begin{tabular}{c|c|rrrr}
				\toprule
				\multirow{2}{*}{Datasets} & \multirow{2}{*}{Methods}  & \multicolumn{4}{c}{Noise Ratio $\eta$} \\ \cline{3-6} 
				
				&  & \multicolumn{1}{c}{0.1}& \multicolumn{1}{c}{0.2} & \multicolumn{1}{c}{0.3} & \multicolumn{1}{c}{0.4} \\ \hline \hline
				\multirow{8}{*}{CIFAR-10} & CCE & 90.69  $\pm$  0.17 & 88.59  $\pm$  0.34 & 86.14  $\pm$  0.40 & 80.11  $\pm$  1.44 \\
				& MAE  & 82.61  $\pm$  4.81 & 52.93  $\pm$  3.60 & 50.36  $\pm$  5.55 & 45.52  $\pm$  0.13 \\
				& Forward $T^\dagger$~\cite{patrini2017making}  & 91.32 $\pm$ 0.21 & 90.35 $\pm$ 0.26  & 89.25 $\pm$ 0.43 & \textbf{88.12 $\pm$ 0.32}\\
				& Forward $\hat{T}~\cite{patrini2017making}$  & 90.52 $\pm$ 0.26 & 89.09 $\pm$ 0.47 & 86.79 $\pm$ 0.36 & 83.55 $\pm$ 0.58 \\
				&Trunc $\mathcal{L}_q$~\cite{zhang2018generalized} &  90.43 $\pm$ 0.25  & 89.45 $\pm$ 0.29 & 87.10 $\pm$ 0.22 & 82.28 $\pm$ 0.67 \\
				\cline{2-6}
				& Baseline (CCE) & 94.31 $\pm$ 0.19& 90.29 $\pm$ 0.35 & 84.61 $\pm$  0.41 & 78.24 $\pm$ 0.82\\
				&Ours &  \textbf{95.69 $\pm$ 0.18}  &  \textbf{94.01 $\pm$  0.22} &  \textbf{ 92.44 $\pm$ 0.37 }& \textbf {85.62 $\pm$ 0.77} \\
				 \hline
				\hline
				\multirow{8}{*}{CIFAR-100} & CCE  &66.54 $\pm$ 0.42 &59.20 $\pm$ 0.18&51.40 $\pm$ 0.16 &42.74 $\pm$ 0.61  \\
				& MAE &13.38 $\pm$ 1.84&11.50 $\pm$ 1.16 & 8.91 $\pm$ 0.89 &8.20 $\pm$ 1.04 \\
				& Forward $T^\dagger$~\cite{patrini2017making} &71.05 $\pm$ 0.30 &{71.08 $\pm$ 0.22}&{70.76 $\pm$ 0.26}&{70.82 $\pm$ 0.45} \\
				& Forward $\hat{T}$~\cite{patrini2017making} &45.96 $\pm$ 1.21 &42.46 $\pm$ 2.16&38.13 $\pm$ 2.97  &34.44 $\pm$ 1.93 \\
				& Trunc $\mathcal{L}_q$~\cite{zhang2018generalized}   &68.86 $\pm$ 0.14&66.59 $\pm$ 0.23 &61.87 $\pm$ 0.39&47.66 $\pm$ 0.69 \\
				\cline{2-6}
				& Baseline (CCE) & 79.40 $\pm$ 0.22 & 73.50 $\pm$ 0.21 & 63.02 $\pm$ 0.32 & 52.06 $\pm$ 0.71\\
				& Ours & \textbf{82.55 $\pm$ 0.24} & \textbf{82.34 $\pm$ 0.20} & \textbf{80.55 $\pm$ 0.26} & \textbf{74.54 $\pm$ 0.64}
				\\ 	\bottomrule
			\end{tabular}
		\end{adjustbox}

	\end{table*}
	
		 	\begin{figure}[t]
	\centering
	\begin{subfigure}{0.5\textwidth}
		\includegraphics[width=0.49\textwidth]{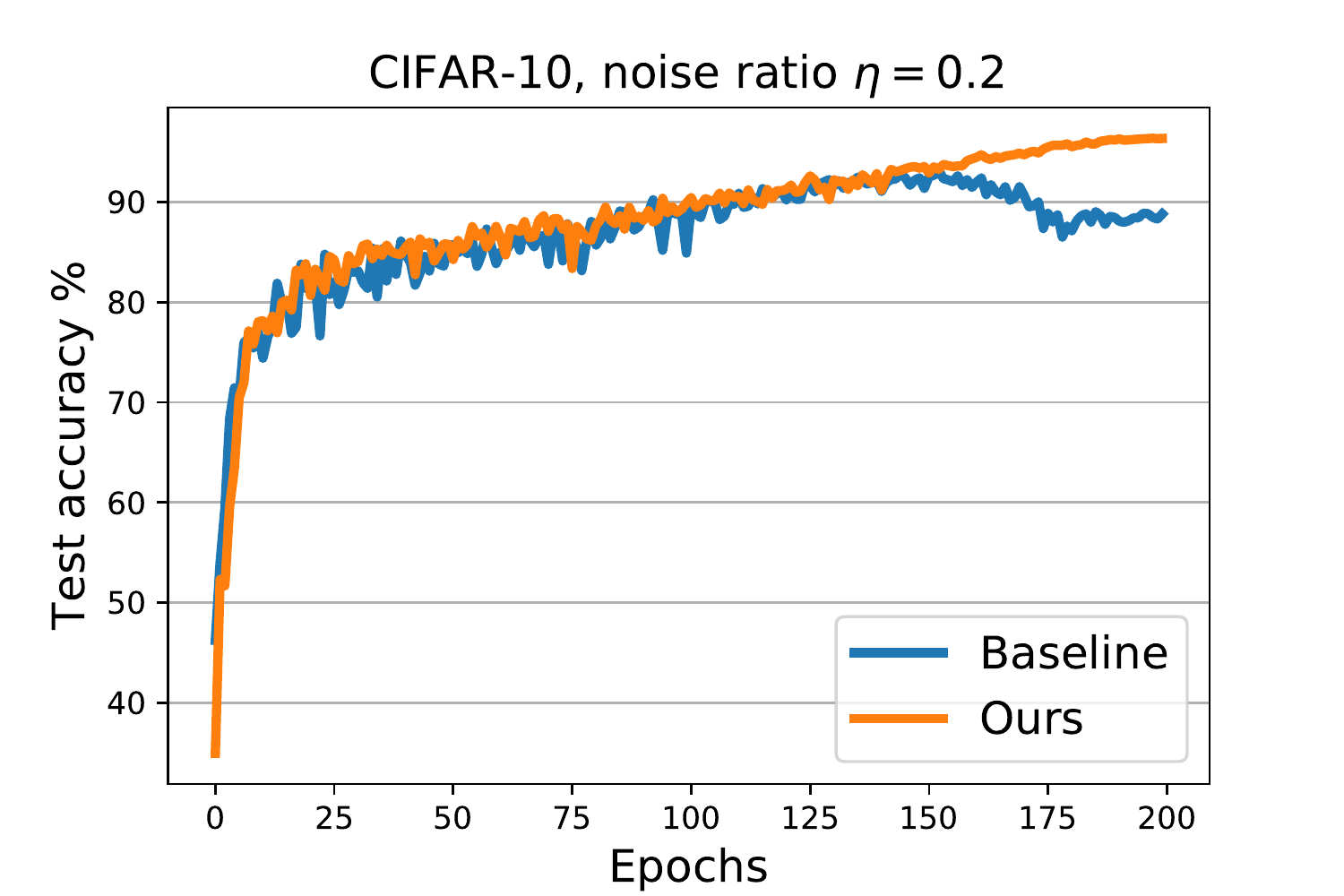}
\includegraphics[width=0.49\textwidth]{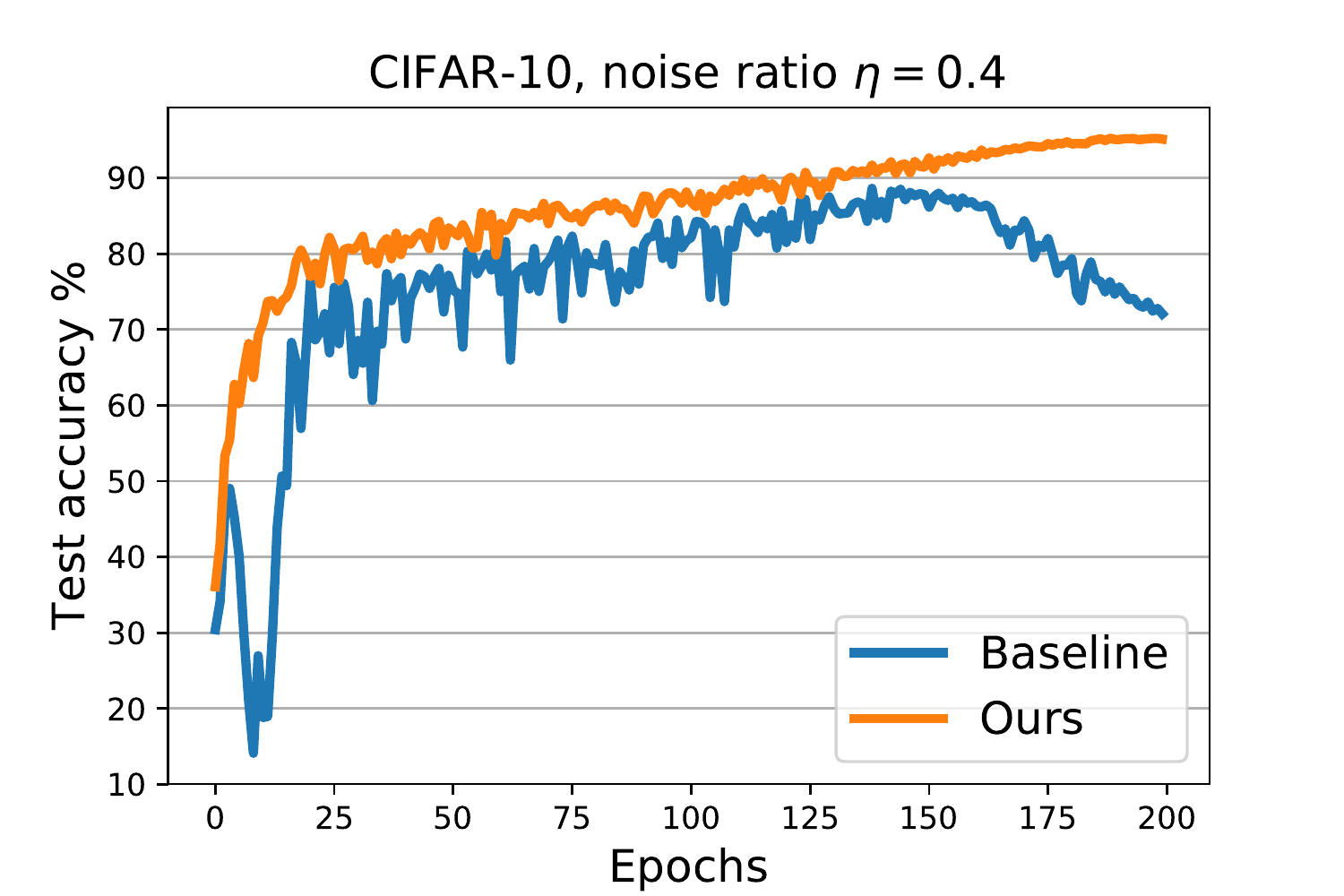}
\end{subfigure} 
	\begin{subfigure}{0.5\textwidth}
		\includegraphics[width=0.49\textwidth]{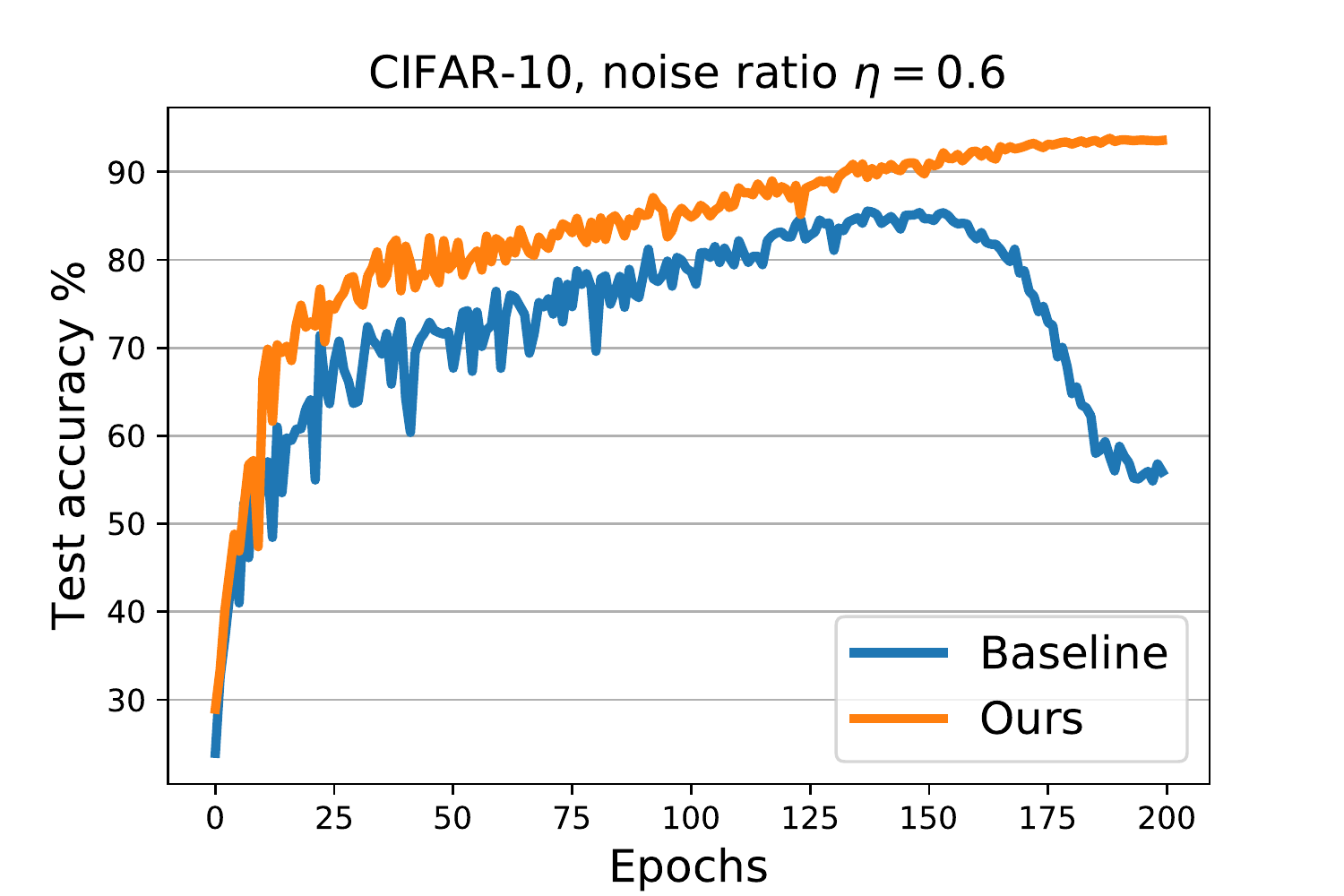}
		\includegraphics[width=0.49\textwidth]{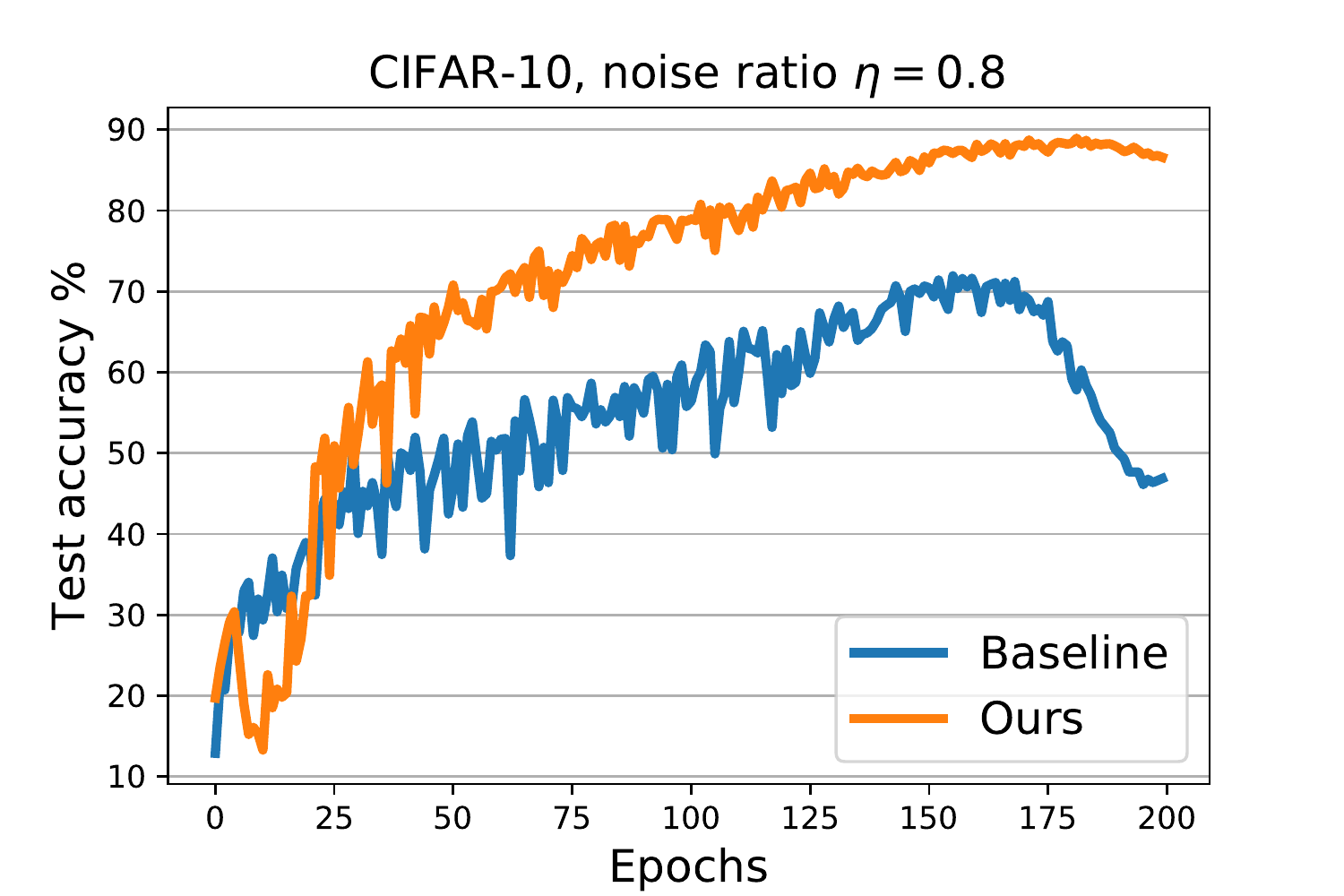}
	\end{subfigure}
	\caption{Test accuracy against the number of epochs on CIFAR-10 under different uniform noise ratio trained with WRN-28-10. Our method is less prone to the label noise over-fitting.}
	\label{testacc}
\end{figure}

	\section{Experiments}
	\label{exp}
	In this section, we present both quantitative and qualitative results to demonstrate the effectiveness of our method.
	Our method is independent of both the architecture and the dataset. Our code is publicly available at \url{https://github.com/xinmei9322/noisylabels}.
	
	\subsection{Experimental setup}

	We first provide results on the widely adopted benchmarks, CIFAR-10 and CIFAR-100. Results on ImageNet and WebVision will be provided in Sec. 5.5 and Sec. 5.6. 
	Following the settings in previous work~\cite{jiang2017mentornet, ren2018learning}, we train wide residual networks WRN-28-10~\cite{zagoruyko2016wide} for 200 epochs with mini-batch size 128. 
All the experiments are trained using momentum 0.9 and weight decay $1\times 10^{-4}$. We use learning rate 0.1 and a cosine annealing schedule as suggested in~\cite{loshchilov2016sgdr}. Our implementation of WRN-28-10 is based on the official code of AutoAugment~\cite{cubuk2018autoaugment}. For the perturbations of $\bxi$, standard data augmentation techniques including random crops and horizontal flips are applied for CIFAR-10 and CIFAR-100.
	
	We use the test error rate and label precision as the evaluation metrics. The label precision is defined as the ratio of the number of selected clean examples to that of total selected examples.
	
	\paragraph{Baselines:} 
	We compare our method with some representative prior works on learning with noisy labels, including:
	\begin{enumerate}
	    \item Bootstrap-hard~\cite{reed2014training}, a self-learning technique that use a convex combination of the given label and model prediction as the training target.
        \item Forward-correction~\cite{patrini2017making}, a loss correction method based on the noise transition matrix $\hat T$ estimated by a pre-trained network.
        \item	D2L~\cite{ma2018dimensionality}, a dimensionality-driven learning strategy which monitors the dimensionality of subspaces and adapts the training target accordingly.		
        \item Generalized Cross Entropy~\cite{zhang2018generalized}, which proposes a class of noise-robust loss functions, \ie $\mathcal{L}_q$ and truncated $\mathcal{L}_q$ loss.
	    \item Co-teaching~\cite{han2018co}, which maintains two networks simultaneously, and cross-trains on instances screened by the ``small loss'' criteria.
        \item MentorNet~\cite{jiang2017mentornet}, a meta-learning model that assigns different weights of training examples based on meta-learned curriculum with clean validation dataset.
        \item Learning to Reweight~\cite{ren2018learning}, an online meta-learning algorithm that learns to assign weights to training examples based on their gradient direction using a clean unbiased validation set.
	
	\end{enumerate}

	\subsection{Input-agnostic uniform label noise}
	First, we test on the uniform random label noise on CIFAR-10 and CIFAR-100.
	Following common practice~\cite{patrini2017making,jiang2017mentornet,zhang2018generalized}, a certain percentage $\eta$ (0\%, 20\%, 40\%, 60\%, 80\%) of true labels on the training dataset are replaced by random labels through uniform sampling. 
	We report the averaged error rates on test datasets over 3 runs. 
	Experimental results are summarized in Table~\ref{cifar} and \ref{cifar100}. 
	Note that different network architectures are used in the competing methods, as pointed out in the table, whose error rates of the base networks are shown in the second column of 0\% noise (clean). The we can observe and compare the relative performance to the standard clean settings. 
	We fix the hyper-parameter $\lambda = 300$ in all the experiments for CIFAR-10 and $\lambda=3000$ for CIFAR-100 following the suggestion in~\cite{laine2016temporal}. Other work~\cite{han2018co,jiang2017mentornet,ren2018learning} selected hyperparameters using an additional clean validation set. If we also have the clean validation set, we can tune $\lambda$ precisely.

	In all the experiments, our method achieves significantly better resistance to label noise from moderate to severe levels. 
	In particular, our approach attains a 13.31\% error rate on CIFAR-10 with a noise fraction of 80\%, down from the previous best 32.08\%. 
	Using the same network architecture WRN-28-10 as ours and 1000 clean validation images, learning to reweight~\cite{ren2018learning} achieves 38.66\% test error on CIFAR-100 with 40\% noise while ours achieves a better 25.73\% even without any knowledge on the clean validation images.
	
\begin{figure}[t]
	\centering\vspace{-0.2cm}
\begin{subfigure}{0.23\textwidth}	
\includegraphics[width=\textwidth]{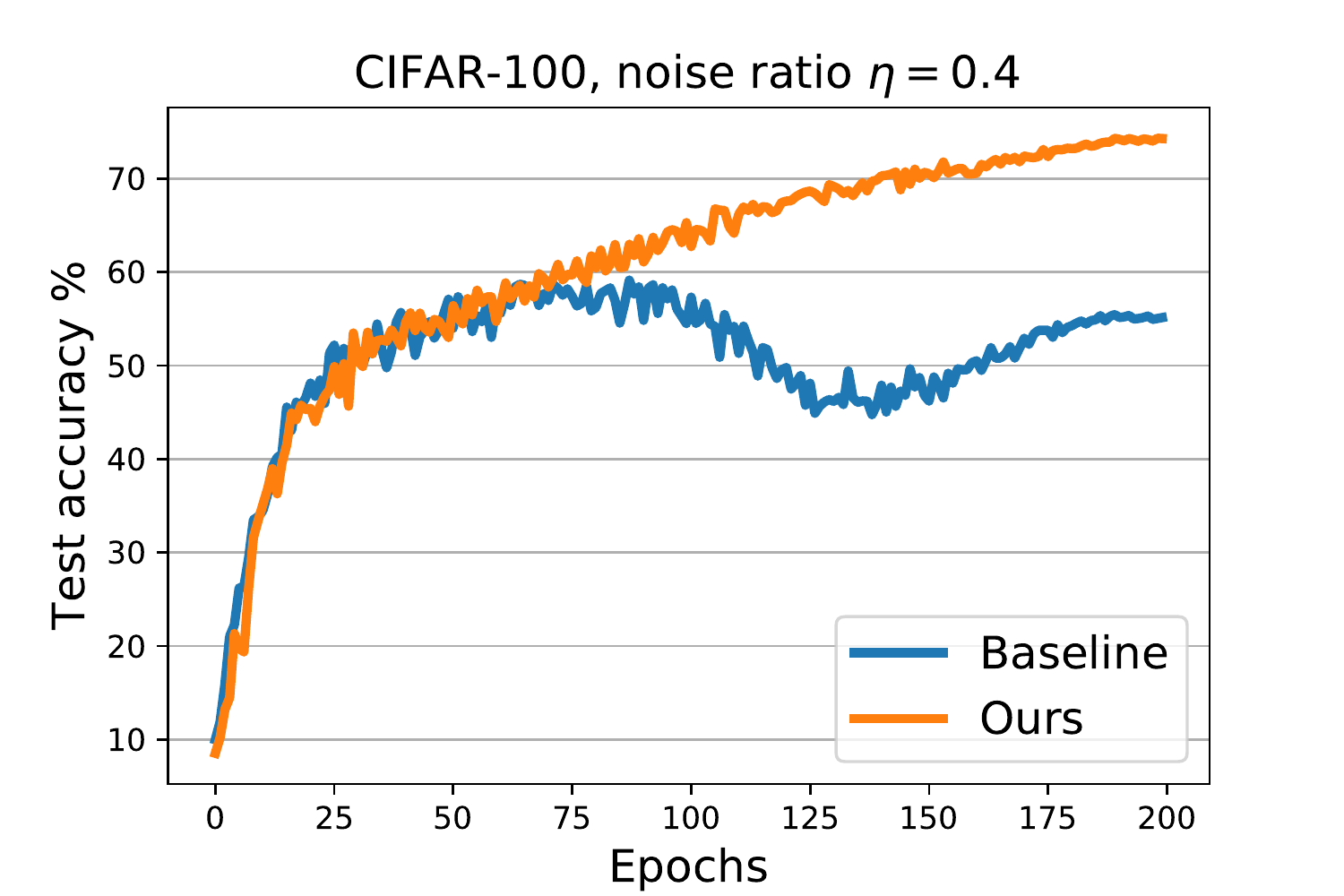}
\end{subfigure} 
	\begin{subfigure}{0.23\textwidth}
		\includegraphics[width=\textwidth]{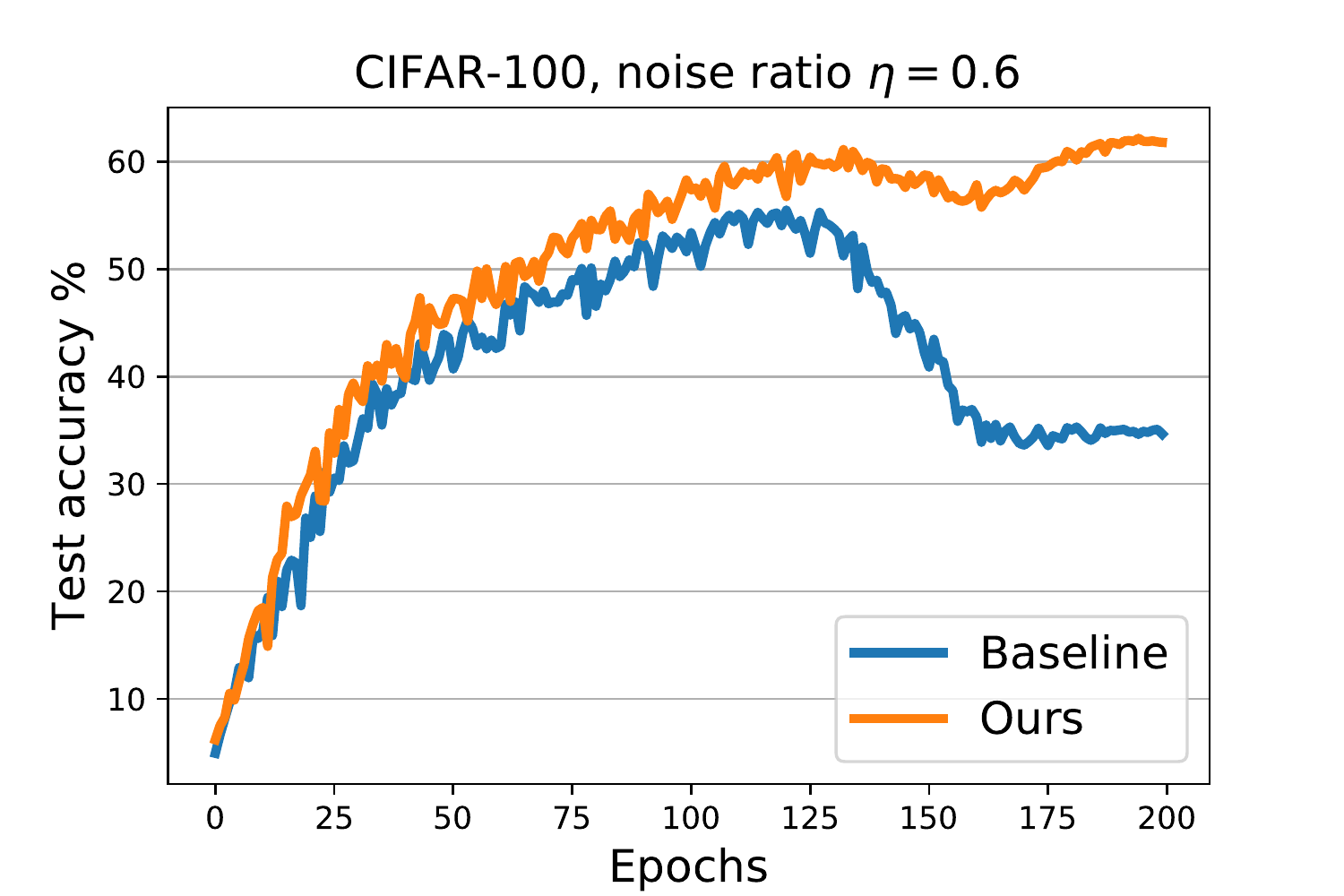}
	\end{subfigure}
	\caption{Test accuracy against the number of epochs on CIFAR-100 under different uniform noise ratios trained with WRN-28-10. Our method is less prone to label noise over-fitting.}
	\label{testacc_cifar100}
\end{figure}

	 Figures~\ref{testacc} and~\ref{testacc_cifar100} plot the test accuracy against the number of epochs on the two datasets. 
	 We provide a baseline -- CCE, standing for Categorical Cross-Entropy loss without the regularizer $\hat R_V$ in Eq.~\ref{equ:2} and train a WRN-28-10. We can see that CCE tends to over-fit the label noise later in the training while our method does not suffer from the incorrect training signals.

\begin{figure}[t]
		\centering\vspace{-0.2cm}
		\begin{subfigure}[b]{0.45\textwidth}
			\includegraphics[width=0.49\textwidth]{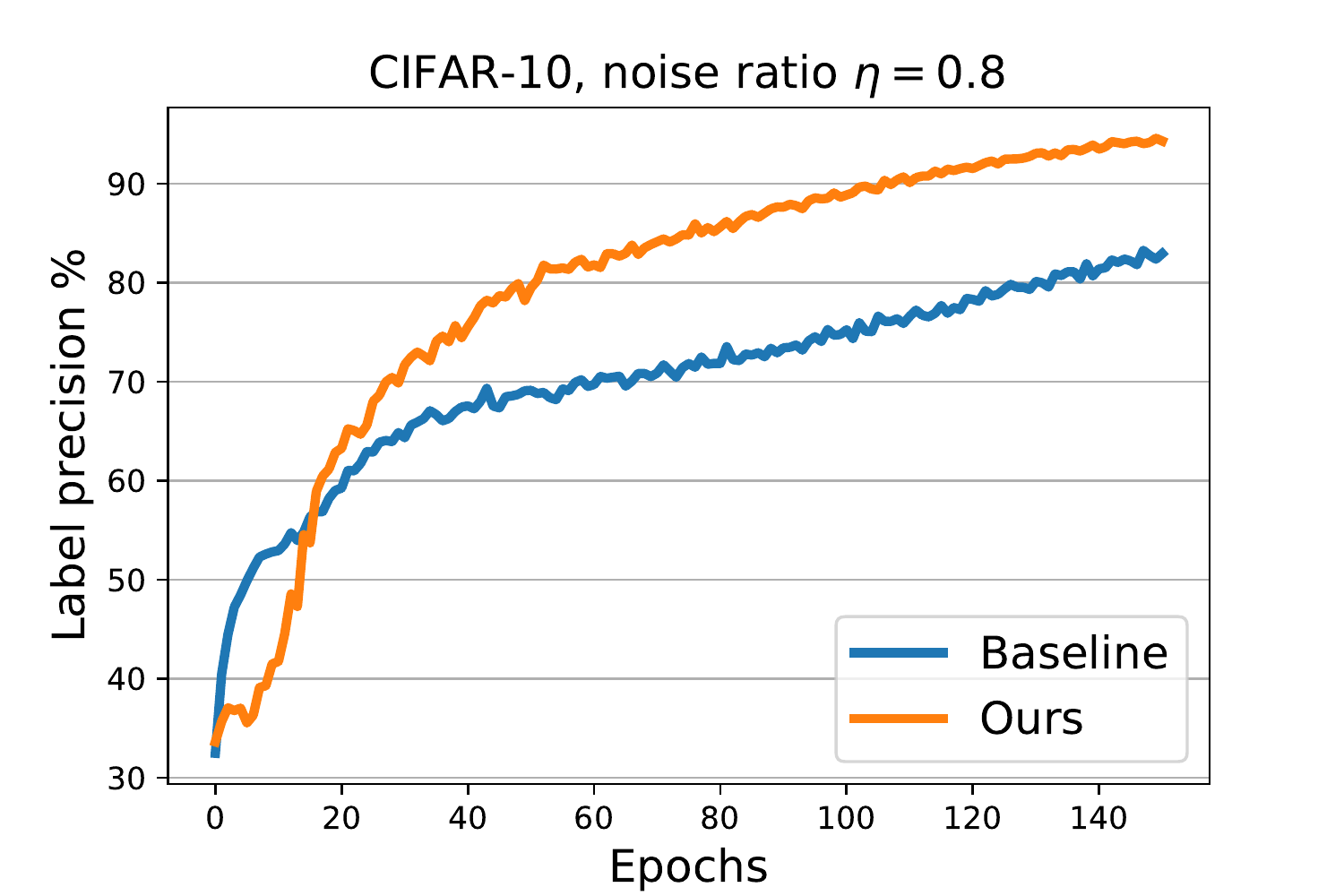}
			\includegraphics[width=0.49\textwidth]{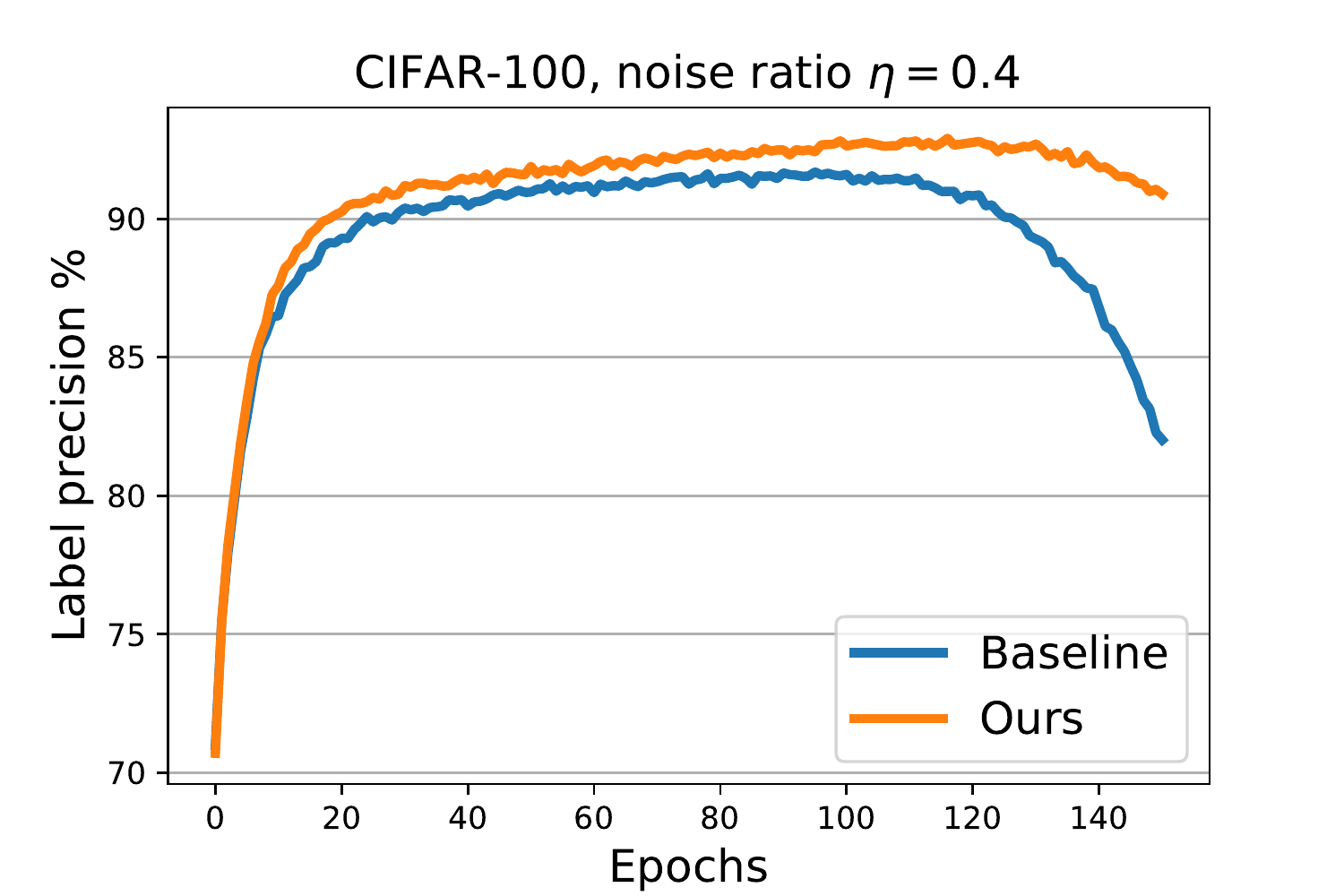}
		
		\end{subfigure}
		\caption{Label precision against the number of epochs on CIFAR-10 (left) and CIFAR-100 (right) with uniform noise, respectively. Here the label precision is computed by the percentage of clean training examples within those having $1-\eta$ minimal training losses. }
		\label{precision}
	\end{figure}

	We also plot the label precision against number of epochs in Figure~\ref{precision}. 
	Here we treat the $1-\eta$ ratio of the training examples with minimal training losses as the candidates. 
	The label precision is computed as the portion of true clean examples among the candidates. 
	An ideal algorithm without any over-fitting will have perfect 100\% label precision. 
	The higher the label precision is, the better robustness the model achieves. Figure~\ref{precision} demonstrates that our method obtains a higher label precision.

	\subsection{Class-dependent asymmetric label noise}
	A more realistic and more challenging noise type than the uniform noise is to corrupt between the semantically similar classes.
	For CIFAR-10, the class-dependent asymmetric noise is simulated by mapping \textsc{truck} $\rightarrow$ \textsc{automobile}, \textsc{bird} $\rightarrow$ \textsc{airplane}, \textsc{deer} $\rightarrow$ \textsc{horse}, \textsc{cat} $\leftrightarrow$ \textsc{dog}, as done in~\cite{patrini2017making, zhang2018generalized}. The noise strength is controlled by the flipping probability $\eta$. The noise transition matrix is:
	\begin{equation}
\scriptsize
 \begin{bmatrix}
    1 & 0 & 0 & 0 & 0 & 0 & 0 & 0 & 0 & 0\\
    0 & 1 & 0 & 0 & 0 & 0 & 0 & 0 & 0 & 0\\
    \eta & 0 & 1-\eta & 0 & 0 & 0 & 0 & 0 & 0 & 0\\
    0 & 0 & 0 & 1-\eta & 0 & \eta & 0 & 0 & 0 & 0\\
    0 & 0 & 0 & 0 & 1-\eta & 0 & 0 & \eta & 0 & 0\\
    0 & 0 & 0 & \eta & 0 & 1-\eta & 0 & 0 & 0 & 0\\
    0 & 0 & 0 & 0 & 0 & 0 & 1 & 0 & 0 & 0\\
    0 & 0 & 0 & 0 & 0 & 0 & 0 & 1 & 0 & 0\\
    0 & 0 & 0 & 0 & 0 & 0 & 0 & 0 & 1 & 0\\
    0 & \eta & 0 & 0 & 0 & 0 & 0 & 0 & 0 & 1-\eta\\
\end{bmatrix}\label{eq:bigT} 
\end{equation}
	For CIFAR-100, class dependent noise is simulated by flipping each class into the next class with probability $\eta$. The last class is flipped to the first class circularly, \ie, the transition matrix has $1-\eta$ on the diagonal and $\eta$ off the diagonal.

\begin{figure}[t]
	\centering
	\begin{subfigure}{0.5\textwidth}
		\includegraphics[width=0.49\textwidth]{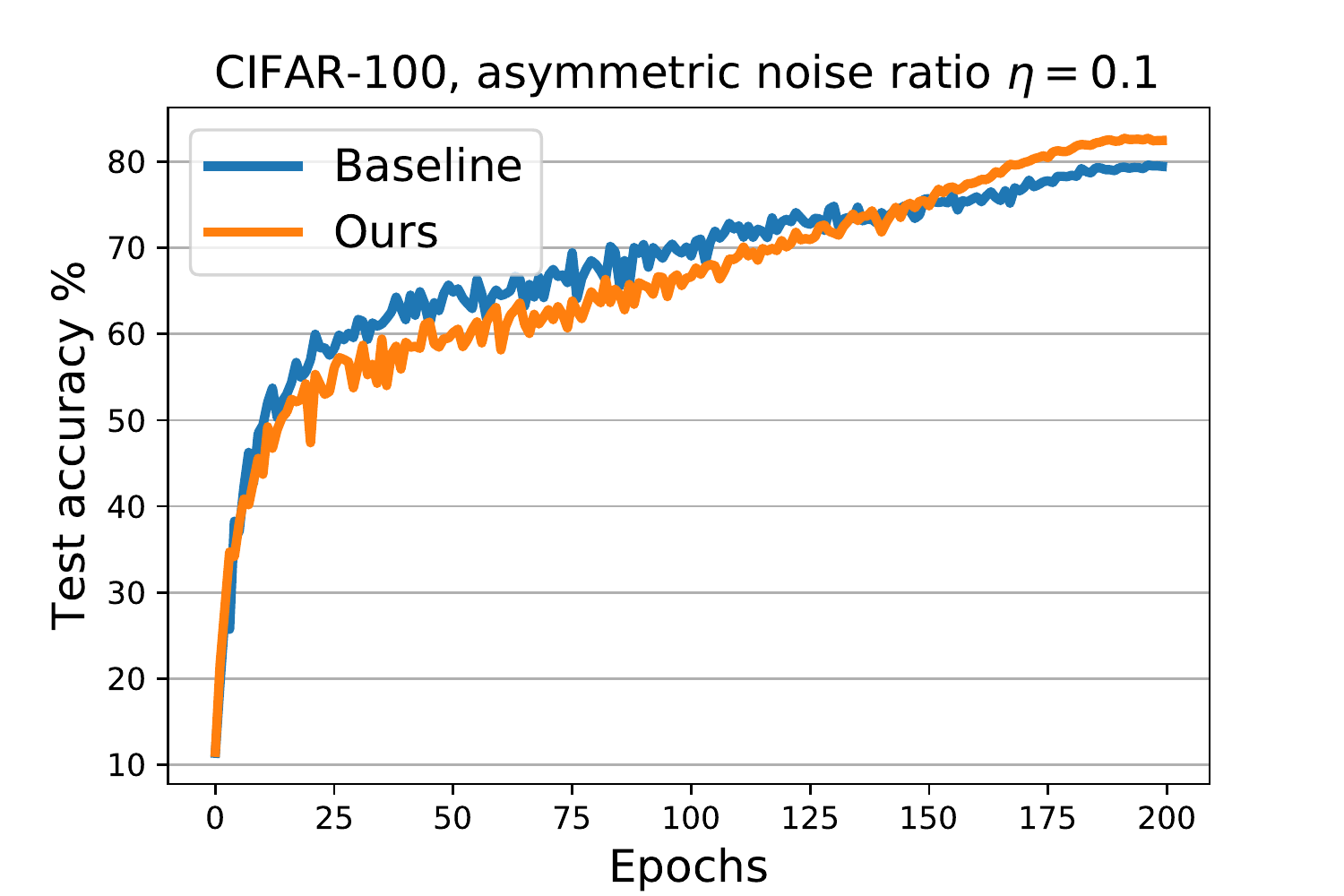}
\includegraphics[width=0.49\textwidth]{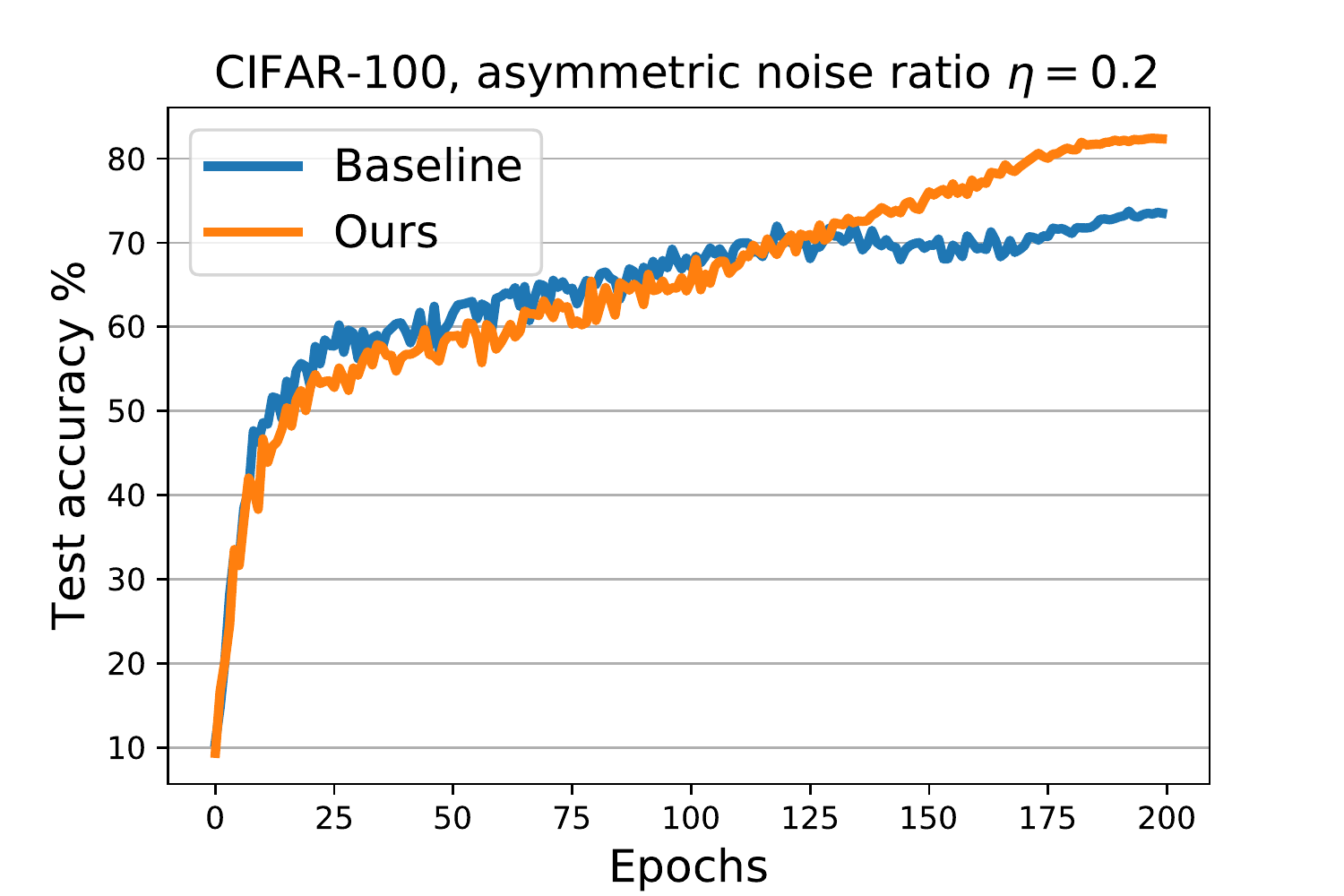}
\end{subfigure} 
	\begin{subfigure}{0.5\textwidth}
		\includegraphics[width=0.49\textwidth]{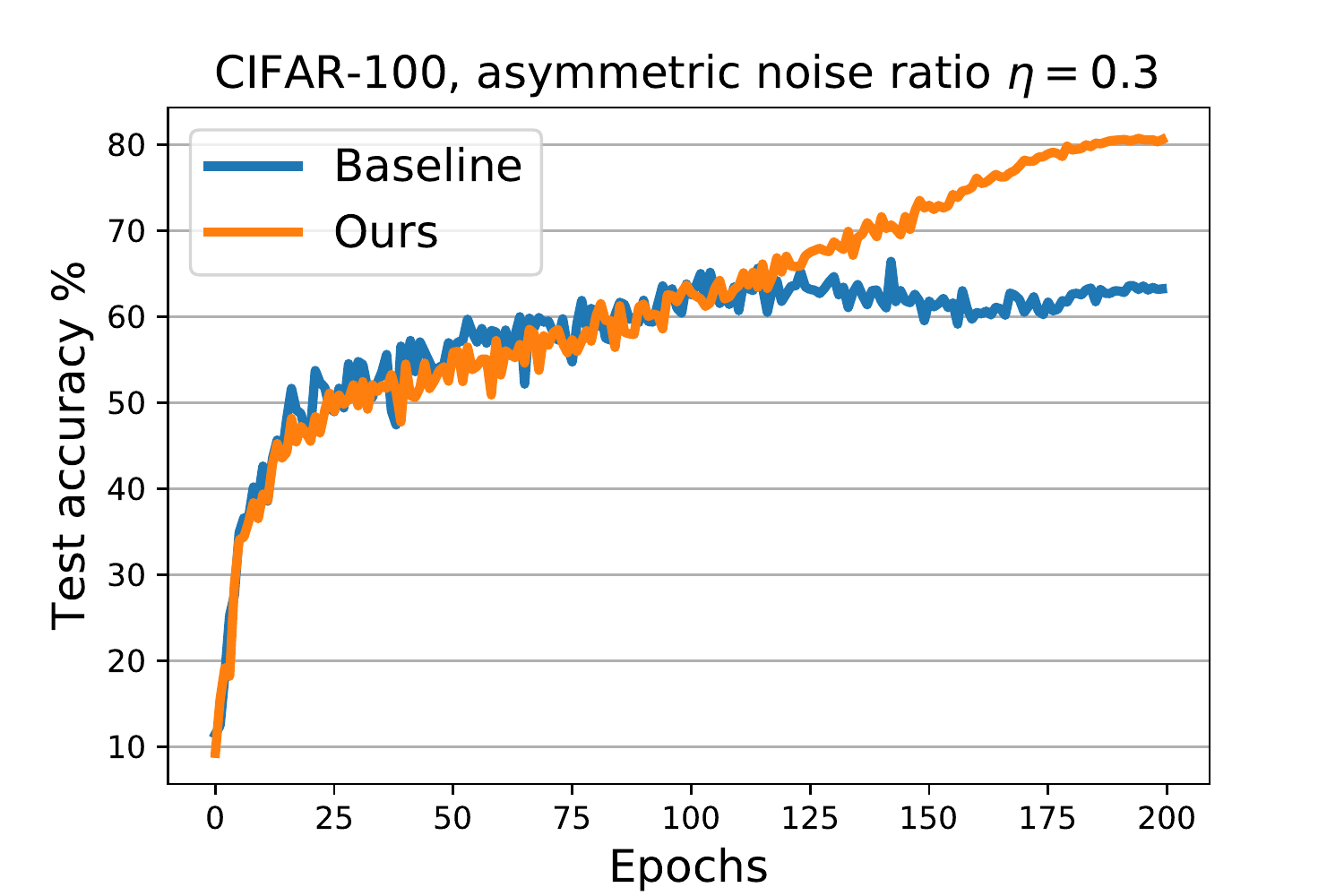}
		\includegraphics[width=0.49\textwidth]{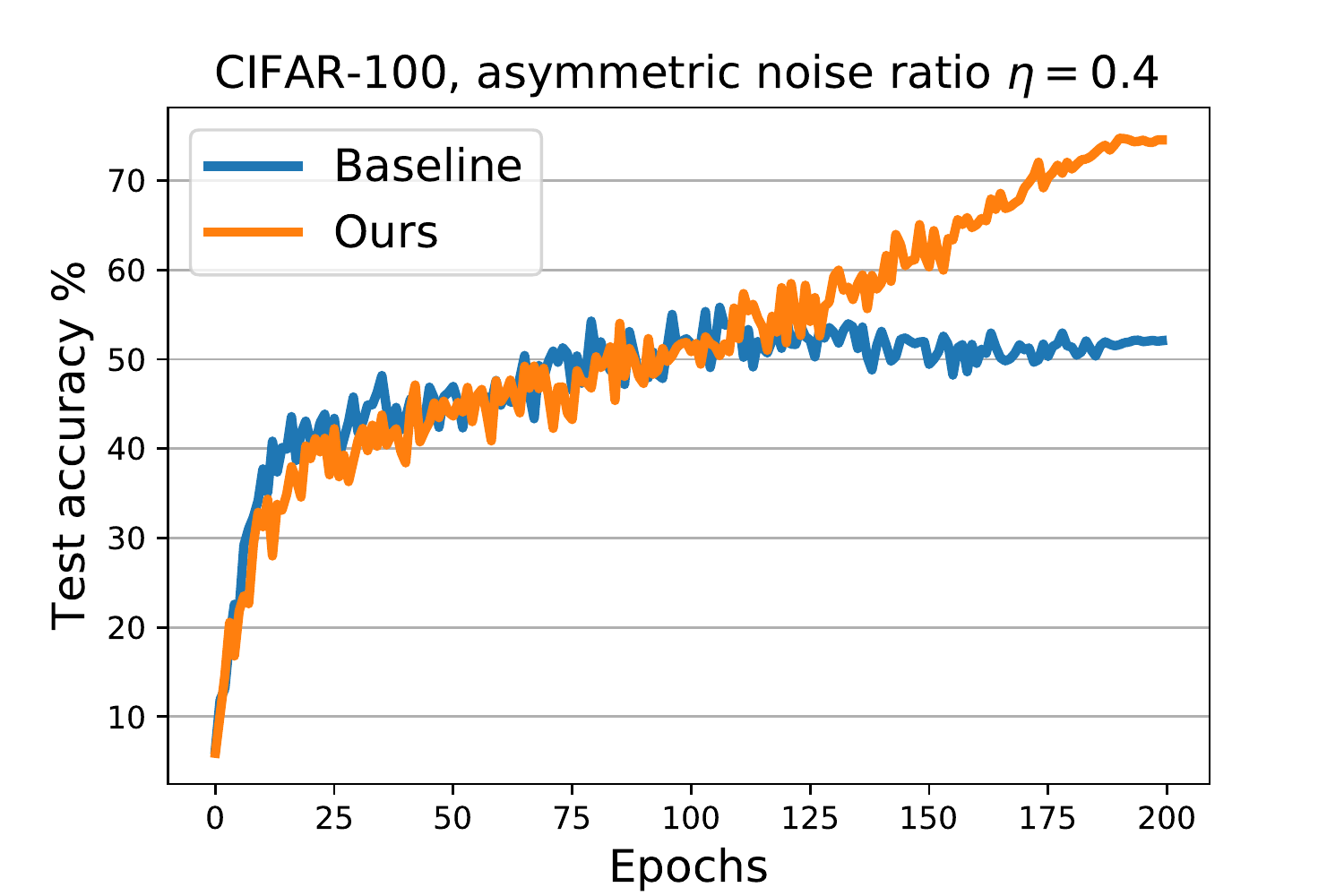}
	\end{subfigure}\vspace{-0.2cm}
	\caption{Test accuracy against the number of epochs on CIFAR-100 under different asymmetric noise ratios trained with WRN-28-10.}
	\label{asytestacc}
\end{figure}

	
	Results are presented in Table~\ref{classdep}. We compare to a range of competing loss-correction methods whose results are taken from~\cite{zhang2018generalized} and the model trained with only CCE without $\hat R_V$. We use the same hyper-parameter $\lambda = 300$ among all the experiments for CIFAR-10 and $\lambda=3000$ for CIFAR-100. Note that Forward $T$ is the forward correction~\cite{patrini2017making} using the ground-truth noise transition matrix, whose results are almost perfect and the comparison is not fair. Our method does not use any ground-truth knowledge of the noise corruption process.
	We can see that our method 
	is robust to all the settings and is less influenced by the variations of noise types.
	The test accuracy along the training process on CIFAR-100 is also plotted in Figure~\ref{asytestacc}.

\subsection{Understanding the model}

\begin{figure}[t]\vspace{-.3cm}
		\centering
			\includegraphics[width=0.3\textwidth]{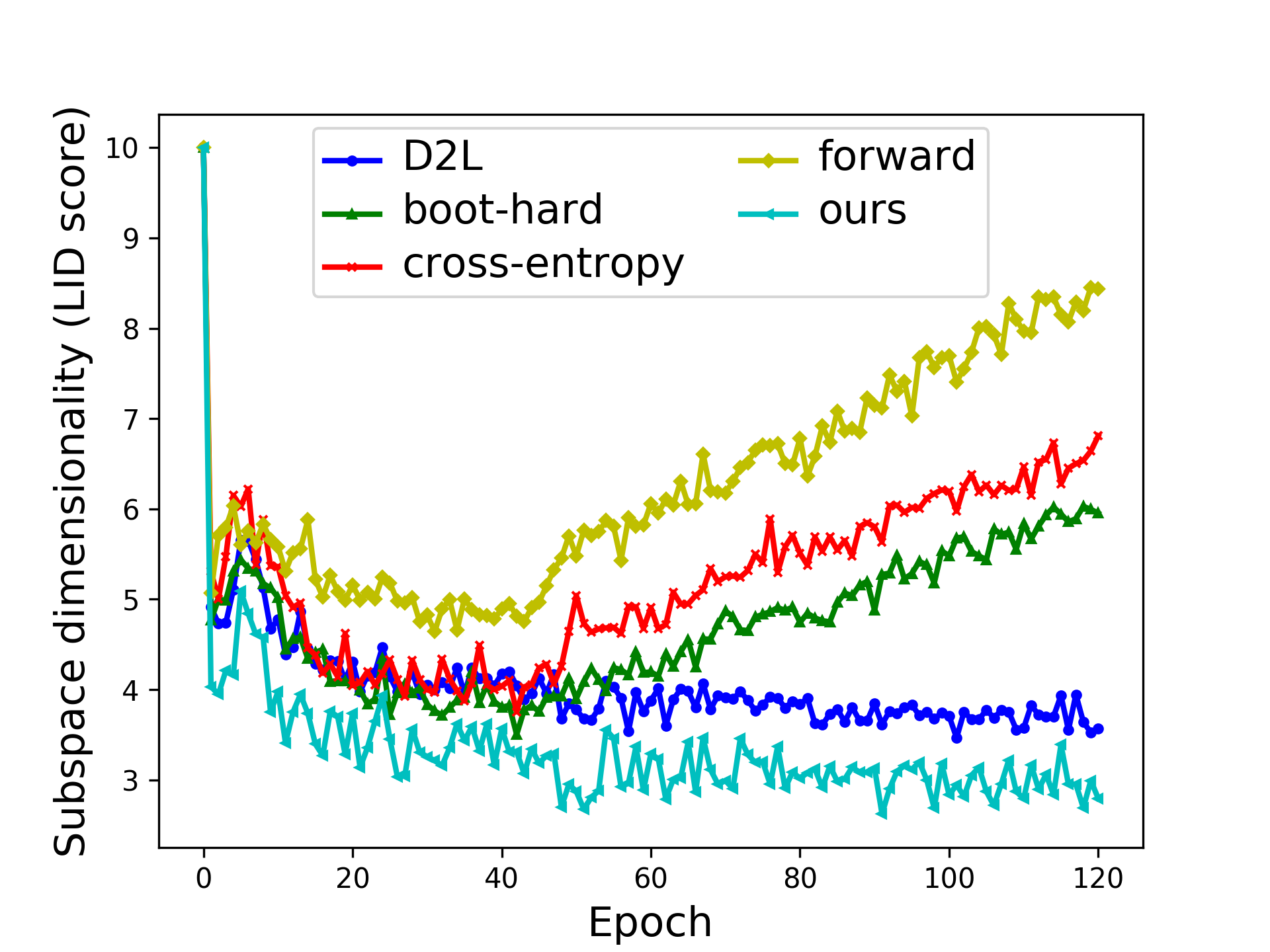}
		\caption{ The estimated subspace dimensionality (averaged LID score). CIFAR-10 with 60\% uniform noise. The dimension of our learned subspace is the lowest.}
		\label{lid}
	\end{figure}

	\textbf{Subspace learning.} 
	We examine the learned subspace dimensionality measured by LID in the training process as described in Section~\ref{sec:lid}. The LID values at the second-to-last layer for CIFAR-10  with 60\% uniform noise are averaged over 10 batches of 128 examples each, for a total of 1280 test examples and the results are shown in Figure~\ref{lid}. We follow the same setup in D2L~\cite{ma2018dimensionality} for fair comparison. It is observed that our method can learn a subspace with lowest dimension, which support the improved generalization performance. Poorly regularized methods like bootstrap-hard and forward-correction tend to decrease LID first and then increase LID to accommodate the noisy examples, which indicates the over-fitting of label noise.
	
\begin{figure}[t]\vspace{-0.2cm}

		\centering
		\includegraphics[width=0.3\textwidth]{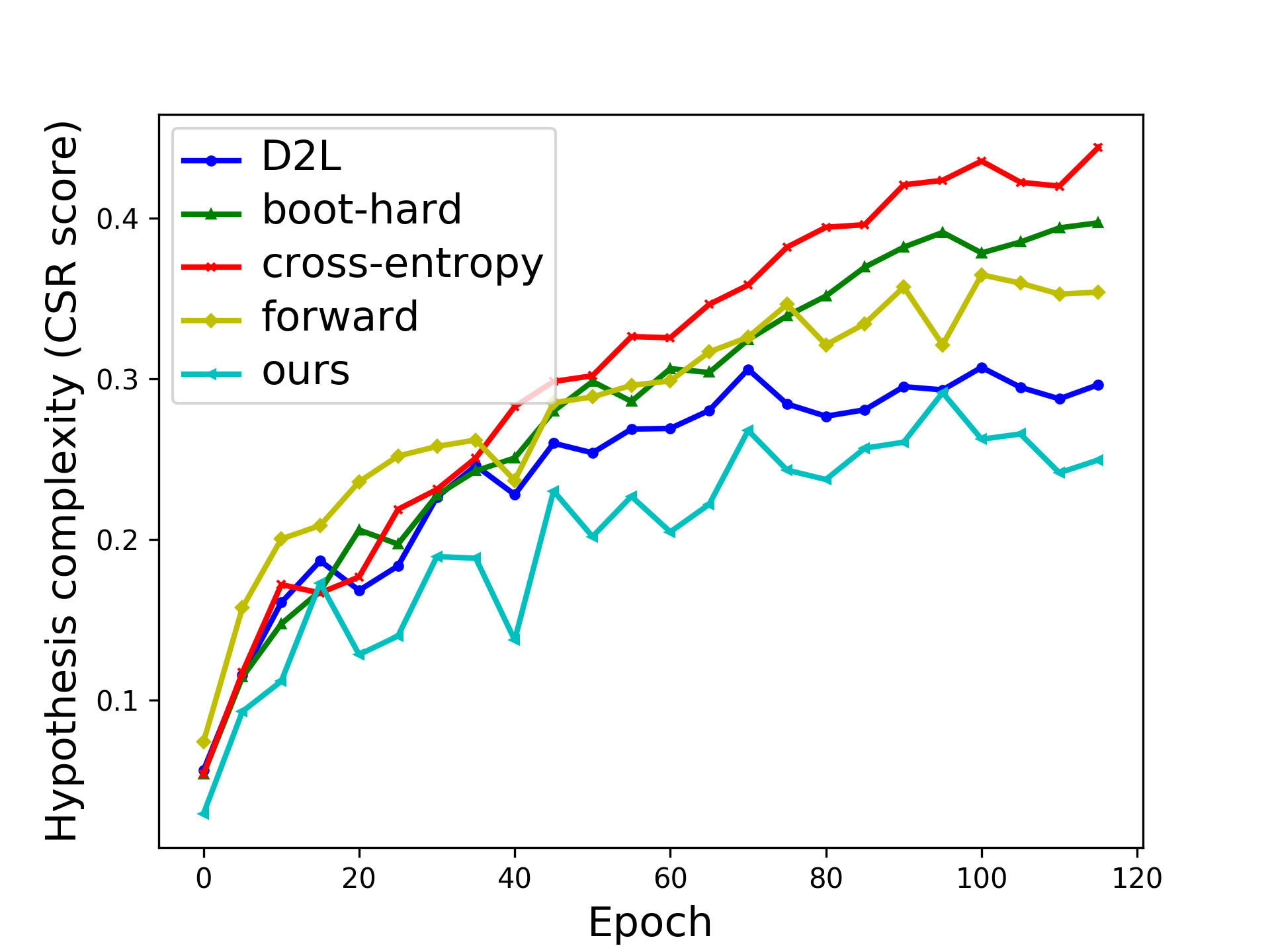}
		\vspace{-0.2cm}
		\caption{The hypothesis complexity measured by CSR. Our method achieved the lowest CSR, which means the samples around decision boundaries are the fewest.}		\vspace{-0.2cm}
		\label{csr}
	\end{figure}

	\textbf{Hypothesis learning.} 
    In order to verify whether a more robust model is learned, we investigate the complexity of the hypothesis. It is known that DNNs with lower complexity is preferable among the models fitting the training data equally well.
    We use the Critical Sample Ratio (CSR)~\cite{arpit2017closer} to measure the hypothesis complexity, which represents the density of data around decision boundaries. We plot the CSR on CIFAR-10 with 60\% uniform noise in Figure~\ref{csr}, showing that our method learns a hypothesis with lower complexity than the other baselines and the samples near the decision boundaries are fewer in our method.
    
	\subsection{Imagenet}
	
	To validate our method for training on large-scale datasets, we test it on ImageNet-2012 classification dataset~\cite{russakovsky2015imagenet}. It contains 1.28 million training images from 1000 classes. We use the 50000 validation images to test the performance. The model is trained under 0\%, 20\%, 40\% uniform label noise. 
	
	We adopt ResNet-50~\cite{he2016deep} as the network and the implementation is based on the official examples of Pytorch.
	 The images are randomly augmented using $10$ degree rotations, random crops with aspect ratio between $3/4$ and $4/3$ and resized to $224\times 224$ pixels, random horizontal flips and color jitters. Finally the images are normalized to have channel-wise zero mean and unit variance on the training set. The hyper-parameter $\lambda$ is set to 50. 
	 We train 90 epochs using cosine schedule with initial learning rate 0.1, momentum 0.9, weight decay $10^{-4}$. The base learning rate is 0.025 with a single cycle of cosine annealing.
	We report top-1 and top-5 validation errors in Table~\ref{imagenet}. We also provide the simple baseline CCE that treats the noisy labels as targets. Our method performs better than MentorNet~\cite{jiang2017mentornet} which uses a data-driven curriculum with inception-resnet v2~\cite{szegedy2017inception}.
	
	\begin{table}[t]\vspace{-0.1cm}
		\caption{Results on the clean Imagenet validation set trained using ResNet-50. Top-1 (Top-5) error rates are listed.}
		\label{imagenet}
	\begin{center}
			\resizebox{0.5\textwidth}{!}{ 	
		\begin{tabular}{l|ccc}
			\toprule
			\multirow{2}{*}{Methods}    &\multicolumn{3}{c}{Noise ratio $\eta$}\\\cline{2-4}
			&0    &0.2   &0.4    \\ 

			\hline\hline
			CCE				& 23.45 (6.78)		&26.41 (8.62)
			& 29.79 (10.58)\\
				\hline
			MentorNet~\cite{jiang2017mentornet}	& -- &--  & 34.9 (14.1)  \\
			\hline
			Ours& \textbf{ 23.27 (6.69)} 		&  \textbf{24.83 (7.74)}	&  \textbf{26.81 (8.99)}  \\
			\bottomrule
		\end{tabular}
	}
	\end{center}
	
	\end{table}
	
	\begin{figure}[t]\vspace{-.2cm}
		\centering
				\begin{subfigure}[b]{0.23\textwidth}
				\includegraphics[width=\textwidth]{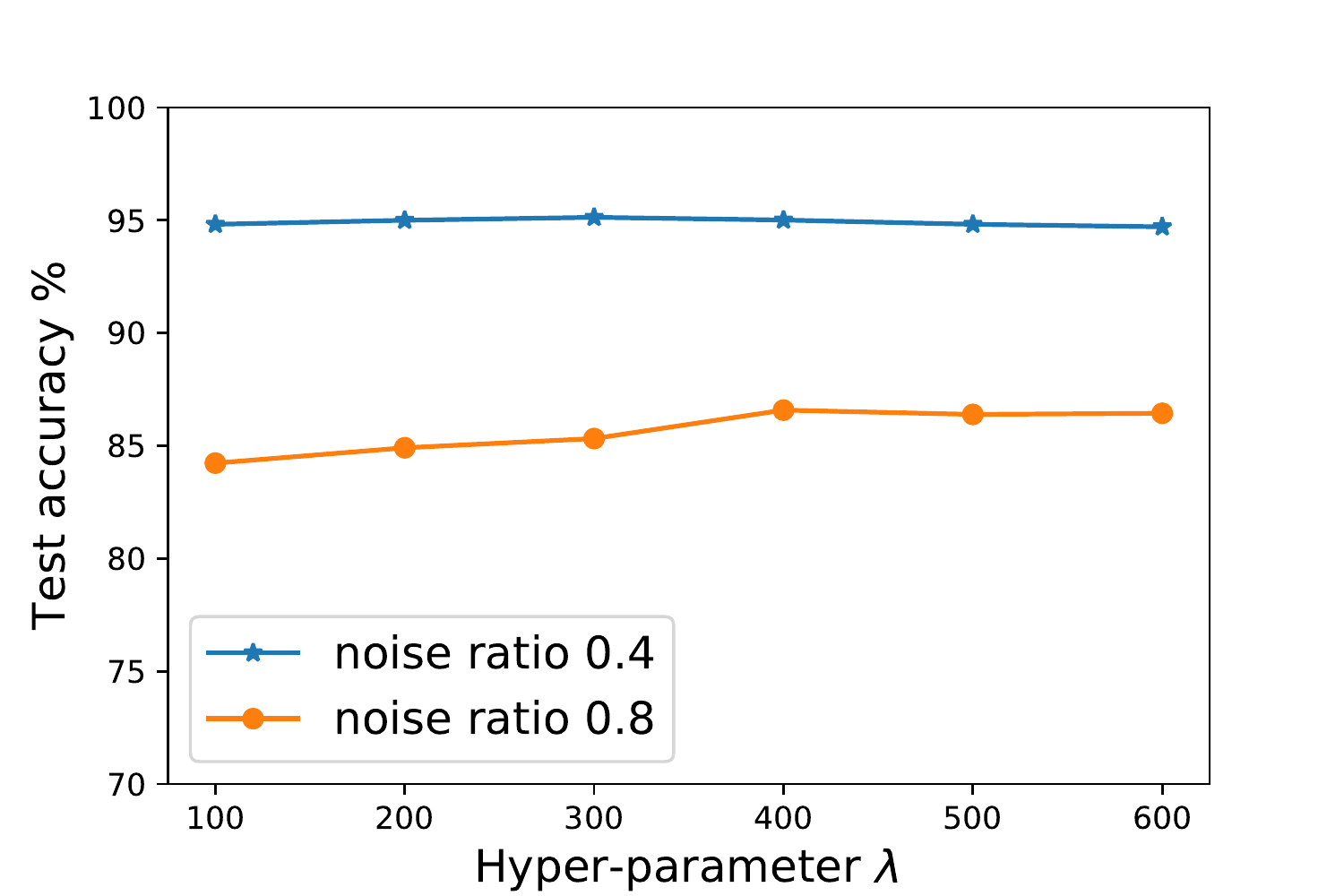}\caption{CIFAR-10} 
			\end{subfigure} 
			\begin{subfigure}[b]{0.23\textwidth}
				\includegraphics[width=\textwidth]{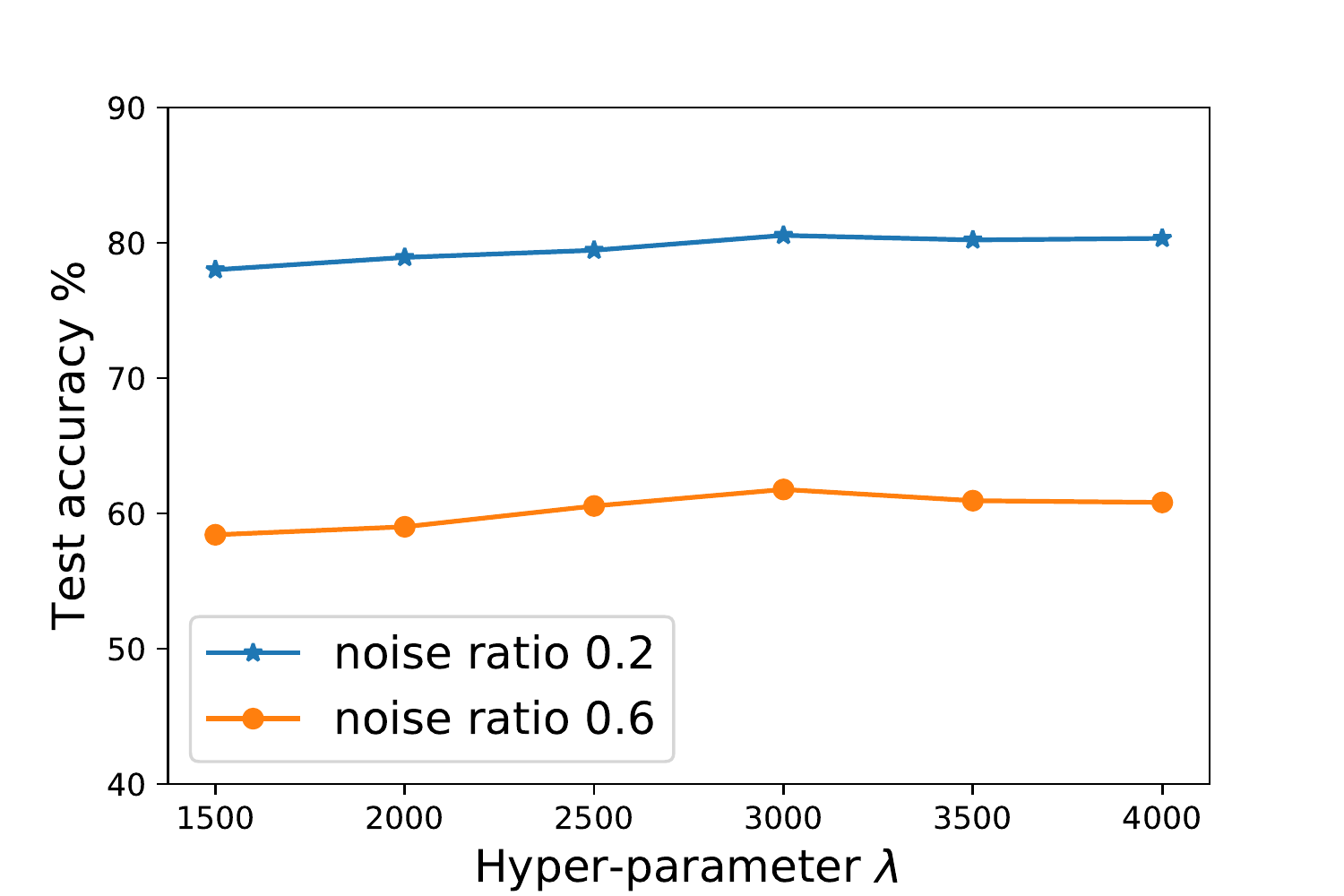}\caption{CIFAR-100} \label{} 
			\end{subfigure}
		\caption{Hyper-parameter sensitivity analysis on CIFAR-10 and CIFAR-100 with various strengths of label noise . Our method is insensitive to a wide range of values for $\lambda$.}
		\label{hyperparam}\vspace{-.2cm}
	\end{figure}
	
	\subsection{Real-world noisy dataset -- WebVision}
	WebVision contains 2.4 million real-world images with noisy labels from Flickr and Google. The 1000 classes are the same as ImageNet. We train the model with ResNet-50~\cite{he2016deep} using the same procedure as the ImageNet experiments. Table~\ref{webvision} shows that our method outperforms previous methods when dealing with a large-scale real noisy dataset. We report the Top-1 and Top-5 error rates on the clean validation datasets of Webvision and Imagenet.
	Note that we do not use additional 30k verification labels from 118 classes.

	\begin{table}[t]\vspace{-.1cm}
				\caption{Results on the clean Webvision validation set and ImageNet validation set. The model is trained on noisy Webvision training data.}
		\label{webvision}
		\begin{center}
			\begin{tabular}{l|cc|cc}
				\toprule
				\multirow{2}{*}{Method}&\multicolumn{2}{c}{Webvision}&\multicolumn{2}{|c}{Imagenet} \\\cline{2-5}
				 & Top-1 & Top-5 & Top-1 & Top-5 \\
				\hline\hline
				Li \etal~\cite{li2017webvision} & 43.0&22.1&52.4&29.6 \\
				Lee \etal~\cite{lee2018cleannet} & 31.5&13.5&39.8&18.9 \\
				MentorNet~\cite{jiang2017mentornet} & 29.2&12.0&37.5&17.0 \\
				Ours &  \textbf{27.3 }& \textbf{10.5 }& \textbf{34.1} &\textbf{ 14.25}\\
				\bottomrule
			\end{tabular}
		\end{center}

	\end{table}

	\subsection{Hyper-parameter sensitivity analysis}
	We assess the sensitivity of our algorithm with respect to the hyper-parameter $\lambda$ and the results are plotted in Figure~\ref{hyperparam}. We can see that the performance of our method remains stable across a wide range of hyper-parameter choices.

	\subsection{Visualization}
	
\begin{figure}[t]\vspace{-0.3cm}
		\centering
		\begin{subfigure}[b]{0.23\textwidth}
			\includegraphics[width=\textwidth]{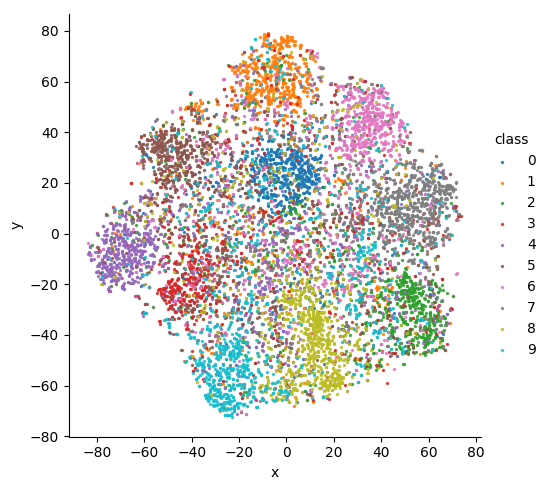}\caption{CCE} 
		\end{subfigure} 
		\begin{subfigure}[b]{0.23\textwidth}
			\includegraphics[width=\textwidth]{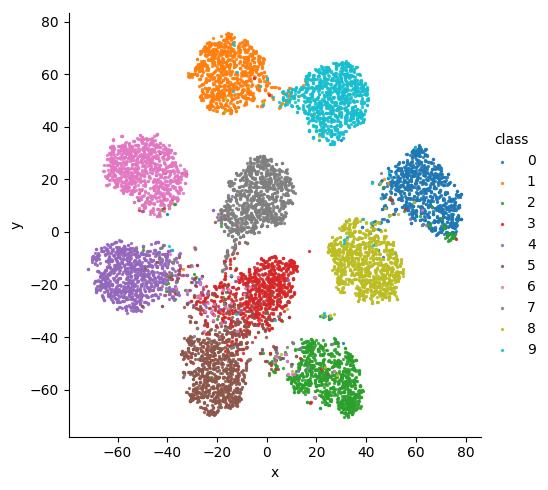}\caption{Ours}
		\end{subfigure} \vspace{-0.2cm}
		\caption{t-SNE 2D embeddings of the test dataset on CIFAR-10 trained with 60\% uniform label noise. Each color represents a class. Our method in (b) learns a more separable feature space than CCE.}
		\label{visual}
	\end{figure}

	We visualize the embedding of our algorithm on test data. 
	 Figure~\ref{visual} shows the representations $h(x)\in \mathbb{R}^{128}$ projected to 2 dimension using t-SNE~\cite{maaten2008visualizing}. It is consistent with the observations of low LID and low CSR in Figure~\ref{lid} and~\ref{csr} that our method learns a low-dimensional subspace where the examples form clusters, far away from the decision boundaries.
	
	\section{Conclusion}
	We propose a simple but effective algorithm for robust deep learning with noisy labels. Our method builds upon a variance-based regularizer that prevents the model from over-fitting to the corrupted labels. We show that the regularizer is an unbiased estimator of Jacobian norm with bounded variance and is closely related to generalization.  Extensive experiments given in the paper show that the generalization performance of DNNs trained with corrupted labels can be improved significantly using our method, which can serve as a strong baseline for deep learning with noisy labels.

	\newpage
	{\small
		\bibliographystyle{ieee}
		\bibliography{egbib}

\begin{thebibliography}{10}\itemsep=-1pt

\bibitem{amodei2016deep}
D.~Amodei, S.~Ananthanarayanan, R.~Anubhai, J.~Bai, E.~Battenberg, C.~Case,
  J.~Casper, B.~Catanzaro, Q.~Cheng, G.~Chen, et~al.
\newblock Deep speech 2: End-to-end speech recognition in english and mandarin.
\newblock In {\em International conference on machine learning}, pages
  173--182, 2016.

\bibitem{arpit2017closer}
D.~Arpit, S.~Jastrzebski, N.~Ballas, D.~Krueger, E.~Bengio, M.~S. Kanwal,
  T.~Maharaj, A.~Fischer, A.~Courville, Y.~Bengio, et~al.
\newblock A closer look at memorization in deep networks.
\newblock {\em arXiv preprint arXiv:1706.05394}, 2017.

\bibitem{azadi2015auxiliary}
S.~Azadi, J.~Feng, S.~Jegelka, and T.~Darrell.
\newblock Auxiliary image regularization for deep cnns with noisy labels.
\newblock {\em arXiv preprint arXiv:1511.07069}, 2015.

\bibitem{cubuk2018autoaugment}
E.~D. Cubuk, B.~Zoph, D.~Mane, V.~Vasudevan, and Q.~V. Le.
\newblock Autoaugment: Learning augmentation policies from data.
\newblock {\em arXiv preprint arXiv:1805.09501}, 2018.

\bibitem{ghosh2017robust}
A.~Ghosh, H.~Kumar, and P.~Sastry.
\newblock Robust loss functions under label noise for deep neural networks.
\newblock In {\em Thirty-First AAAI Conference on Artificial Intelligence},
  2017.

\bibitem{goldberger2016training}
J.~Goldberger and E.~Ben-Reuven.
\newblock Training deep neural-networks using a noise adaptation layer.
\newblock {\em ICLR}, 2017.

\bibitem{han2018co}
B.~Han, Q.~Yao, X.~Yu, G.~Niu, M.~Xu, W.~Hu, I.~Tsang, and M.~Sugiyama.
\newblock Co-teaching: Robust training of deep neural networks with extremely
  noisy labels.
\newblock In {\em Advances in Neural Information Processing Systems}, pages
  8536--8546, 2018.

\bibitem{he2016deep}
K.~He, X.~Zhang, S.~Ren, and J.~Sun.
\newblock Deep residual learning for image recognition.
\newblock In {\em Proceedings of the IEEE conference on computer vision and
  pattern recognition}, pages 770--778, 2016.

\bibitem{hendrycks2018using}
D.~Hendrycks, M.~Mazeika, D.~Wilson, and K.~Gimpel.
\newblock Using trusted data to train deep networks on labels corrupted by
  severe noise.
\newblock {\em arXiv preprint arXiv:1802.05300}, 2018.

\bibitem{houle2017local}
M.~E. Houle.
\newblock Local intrinsic dimensionality i: an extreme-value-theoretic
  foundation for similarity applications.
\newblock In {\em International Conference on Similarity Search and
  Applications}, pages 64--79. Springer, 2017.

\bibitem{jean2018semi}
N.~Jean, S.~M. Xie, and S.~Ermon.
\newblock Semi-supervised deep kernel learning: Regression with unlabeled data
  by minimizing predictive variance.
\newblock In {\em Advances in Neural Information Processing Systems}, pages
  5327--5338, 2018.

\bibitem{jiang2017mentornet}
L.~Jiang, Z.~Zhou, T.~Leung, L.-J. Li, and L.~Fei-Fei.
\newblock Mentornet: Learning data-driven curriculum for very deep neural
  networks on corrupted labels.
\newblock {\em arXiv preprint arXiv:1712.05055}, 2017.

\bibitem{laine2016temporal}
S.~Laine and T.~Aila.
\newblock Temporal ensembling for semi-supervised learning.
\newblock {\em arXiv preprint arXiv:1610.02242}, 2016.

\bibitem{lee2018cleannet}
K.-H. Lee, X.~He, L.~Zhang, and L.~Yang.
\newblock Cleannet: Transfer learning for scalable image classifier training
  with label noise.
\newblock In {\em Proceedings of the IEEE Conference on Computer Vision and
  Pattern Recognition}, pages 5447--5456, 2018.

\bibitem{li2017webvision}
W.~Li, L.~Wang, W.~Li, E.~Agustsson, and L.~Van~Gool.
\newblock Webvision database: Visual learning and understanding from web data.
\newblock {\em arXiv preprint arXiv:1708.02862}, 2017.

\bibitem{loshchilov2016sgdr}
I.~Loshchilov and F.~Hutter.
\newblock Sgdr: Stochastic gradient descent with warm restarts.
\newblock {\em arXiv preprint arXiv:1608.03983}, 2016.

\bibitem{ma2018dimensionality}
X.~Ma, Y.~Wang, M.~E. Houle, S.~Zhou, S.~M. Erfani, S.-T. Xia, S.~Wijewickrema,
  and J.~Bailey.
\newblock Dimensionality-driven learning with noisy labels.
\newblock In {\em International Conference on Machine Learning}, pages
  3361--3370, 2018.

\bibitem{maaten2008visualizing}
L.~v.~d. Maaten and G.~Hinton.
\newblock Visualizing data using t-sne.
\newblock {\em Journal of Machine Learning Research}, 9(Nov):2579--2605, 2008.

\bibitem{menon2016learning}
A.~K. Menon, B.~Van~Rooyen, and N.~Natarajan.
\newblock Learning from binary labels with instance-dependent corruption.
\newblock {\em arXiv preprint arXiv:1605.00751}, 2016.

\bibitem{natarajan2013learning}
N.~Natarajan, I.~S. Dhillon, P.~K. Ravikumar, and A.~Tewari.
\newblock Learning with noisy labels.
\newblock In {\em Advances in neural information processing systems}, pages
  1196--1204, 2013.

\bibitem{novak2018sensitivity}
R.~Novak, Y.~Bahri, D.~A. Abolafia, J.~Pennington, and J.~Sohl-Dickstein.
\newblock Sensitivity and generalization in neural networks: an empirical
  study.
\newblock {\em arXiv preprint arXiv:1802.08760}, 2018.

\bibitem{patrini2017making}
G.~Patrini, A.~Rozza, A.~Krishna~Menon, R.~Nock, and L.~Qu.
\newblock Making deep neural networks robust to label noise: A loss correction
  approach.
\newblock In {\em Proceedings of the IEEE Conference on Computer Vision and
  Pattern Recognition}, pages 1944--1952, 2017.

\bibitem{reed2014training}
S.~Reed, H.~Lee, D.~Anguelov, C.~Szegedy, D.~Erhan, and A.~Rabinovich.
\newblock Training deep neural networks on noisy labels with bootstrapping.
\newblock {\em arXiv preprint arXiv:1412.6596}, 2014.

\bibitem{ren2018learning}
M.~Ren, W.~Zeng, B.~Yang, and R.~Urtasun.
\newblock Learning to reweight examples for robust deep learning.
\newblock {\em arXiv preprint arXiv:1803.09050}, 2018.

\bibitem{ren2015faster}
S.~Ren, K.~He, R.~Girshick, and J.~Sun.
\newblock Faster r-cnn: Towards real-time object detection with region proposal
  networks.
\newblock In {\em Advances in neural information processing systems}, pages
  91--99, 2015.

\bibitem{russakovsky2015imagenet}
O.~Russakovsky, J.~Deng, H.~Su, J.~Krause, S.~Satheesh, S.~Ma, Z.~Huang,
  A.~Karpathy, A.~Khosla, M.~Bernstein, et~al.
\newblock Imagenet large scale visual recognition challenge.
\newblock {\em International journal of computer vision}, 115(3):211--252,
  2015.

\bibitem{shwartz2017opening}
R.~Shwartz-Ziv and N.~Tishby.
\newblock Opening the black box of deep neural networks via information.
\newblock {\em arXiv preprint arXiv:1703.00810}, 2017.

\bibitem{sokolic2017robust}
J.~Sokoli{\'c}, R.~Giryes, G.~Sapiro, and M.~R. Rodrigues.
\newblock Robust large margin deep neural networks.
\newblock {\em IEEE Transactions on Signal Processing}, 65(16):4265--4280,
  2017.

\bibitem{szegedy2017inception}
C.~Szegedy, S.~Ioffe, V.~Vanhoucke, and A.~A. Alemi.
\newblock Inception-v4, inception-resnet and the impact of residual connections
  on learning.
\newblock In {\em Thirty-First AAAI Conference on Artificial Intelligence},
  2017.

\bibitem{tarvainen2017mean}
A.~Tarvainen and H.~Valpola.
\newblock Mean teachers are better role models: Weight-averaged consistency
  targets improve semi-supervised deep learning results.
\newblock In {\em Advances in neural information processing systems}, pages
  1195--1204, 2017.

\bibitem{vahdat2017toward}
A.~Vahdat.
\newblock Toward robustness against label noise in training deep discriminative
  neural networks.
\newblock In {\em Advances in Neural Information Processing Systems}, pages
  5596--5605, 2017.

\bibitem{xiao2015learning}
T.~Xiao, T.~Xia, Y.~Yang, C.~Huang, and X.~Wang.
\newblock Learning from massive noisy labeled data for image classification.
\newblock In {\em Proceedings of the IEEE Conference on Computer Vision and
  Pattern Recognition}, pages 2691--2699, 2015.

\bibitem{zagoruyko2016wide}
S.~Zagoruyko and N.~Komodakis.
\newblock Wide residual networks.
\newblock {\em arXiv preprint arXiv:1605.07146}, 2016.

\bibitem{zhang2016understanding}
C.~Zhang, S.~Bengio, M.~Hardt, B.~Recht, and O.~Vinyals.
\newblock Understanding deep learning requires rethinking generalization.
\newblock {\em arXiv preprint arXiv:1611.03530}, 2016.

\bibitem{zhang2018generalized}
Z.~Zhang and M.~Sabuncu.
\newblock Generalized cross entropy loss for training deep neural networks with
  noisy labels.
\newblock In {\em Advances in Neural Information Processing Systems}, pages
  8792--8802, 2018.

\bibitem{zhu2014bayesian}
J.~Zhu, N.~Chen, and E.~P. Xing.
\newblock Bayesian inference with posterior regularization and applications to
  infinite latent svms.
\newblock {\em The Journal of Machine Learning Research}, 15(1):1799--1847,
  2014.

\end{thebibliography}
	}
	
	\newpage
	
	\appendix
	\section{Experimental details}	

The implementation of experiments on CIFAR-10 and CIFAR-100 is based on the official code in~\cite{cubuk2018autoaugment}. We maintain the default settings of WRN-28-10, which is also comparable to Learning to reweight~\cite{ren2018learning}. Since we would like to provide a simple but effective baseline in robust deep learning, we do not tune the hyperparameter $\lambda$ for each experiments, but fix the same $\lambda =300$ for all the experiments on CIFAR-10 and $\lambda =3000$ for CIFAR-100. 
And another technique is that we ramp-up with sigmoid the regularization from $0$ to $\lambda$ in the beginning of the training, \ie, the first $5$ epochs. It is reasonable because of the initial low prediction accuracy and it is a common practice in~\cite{laine2016temporal,tarvainen2017mean}. For the experiments on ImageNet and WebVision, we largely follow the default setting of Pytorch official examples.
And for the analysis experiments, we make use of the source code provided by D2L~\cite{ma2018dimensionality}. In order to have a fair comparison, we adapt our method to their framework and maintain the same epochs and other settings as the competing baselines.

\end{document}